%% file: main.tex
\documentclass[10pt,letterpaper]{article}
\usepackage[margin=1in]{geometry}
\usepackage{iftex}
\usepackage{amsmath,amssymb,textcomp}
\usepackage{graphicx}
\usepackage[table]{xcolor}
\usepackage{booktabs,multirow,array,float,longtable}
\usepackage{algorithm,algpseudocode}
\usepackage{cite}
\usepackage{setspace}
\usepackage[aboveskip=1pt,font=small,labelfont=bf,labelsep=period,justification=raggedright,singlelinecheck=off]{caption}
\usepackage{titlesec}
\usepackage[hidelinks]{hyperref}
\ifPDFTeX
\usepackage[strings]{underscore}
\fi

\DeclareCaptionLabelFormat{supplementary}{#2~#1}
\makeatletter
\renewcommand{\@biblabel}[1]{\quad#1.}
\makeatother
\newcolumntype{L}{@{\extracolsep{\fill}}l}
\newcolumntype{C}{@{\extracolsep{\fill}}c}
\newcolumntype{R}{@{\extracolsep{\fill}}r}
\newcommand{\tblwidth}{\linewidth}
\DeclareMathOperator*{\argmin}{arg\,min}
\DeclareMathOperator*{\argmax}{arg\,max}
\DeclareMathOperator{\LN}{LN}
\newcommand{\Eqref}[1]{Eq~(\ref{#1})}
\raggedright
\hypersetup{pdftitle={Agentic neural architecture search for clinical prediction from electronic health records: Cross-hospital knowledge transfer and computational efficiency},pdfauthor={Deyi Li, Qi Xu, Lingyao Li, Tiansheng Wang, Muxuan Liang, Mei Liu}}
\title{Agentic neural architecture search for clinical prediction from electronic health records: Cross-hospital knowledge transfer and computational efficiency}
\author{Deyi Li$^{1,\dagger,*}$, Qi Xu$^{1,\dagger}$, Lingyao Li$^{2}$\\
Tiansheng Wang$^{3,4}$, Muxuan Liang$^{5}$, Mei Liu$^{1,*}$}
\date{}
\begin{document}
\maketitle
\begingroup
\small
\noindent\textsuperscript{1} Department of Health Outcomes and Biomedical Informatics, College of Medicine, University of Florida, Gainesville, FL, USA\par
\noindent\textsuperscript{2} College of Information Science, University of Arizona, Tucson, AZ, USA\par
\noindent\textsuperscript{3} Department of Pharmaceutical Health Outcomes and Policy, College of Pharmacy, University of Houston, Houston, TX, USA\par
\noindent\textsuperscript{4} Department of Epidemiology, Gillings School of Global Public Health, University of North Carolina at Chapel Hill, Chapel Hill, NC, USA\par
\noindent\textsuperscript{5} Department of Biostatistics, University of Texas MD Anderson Cancer Center, Houston, TX, USA\par
\smallskip
\noindent$^{\dagger}$ These authors contributed equally to this work.\par
\noindent$^{*}$ Corresponding authors: lideyi@ufl.edu (DL); mei.liu@ufl.edu (ML).\par
\noindent ORCID: Deyi Li, 0009-0004-7039-7303; Mei Liu, 0000-0002-8036-2110.\par
\endgroup
\medskip
\section*{Abstract}
Transformer-based models are widely used for clinical prediction from electronic health records (EHRs), yet their architectures require manual tuning, and the optimal configuration may vary across tasks and hospitals. Neural architecture search (NAS) automates architecture design, but conventional methods are computationally costly for Transformer-based EHR models. Recent large language model (LLM)-guided NAS methods reduce manual search design but conduct each search independently, without reusing architecture knowledge across hospitals. In this study, we propose ATHENA (\textbf{A}gentic \textbf{T}ransfer across \textbf{H}ospitals for \textbf{E}HR \textbf{N}eural \textbf{A}rchitecture Search), a knowledge-guided agentic NAS framework for Transformer-based EHR modeling. ATHENA uses a weight-sharing supernet that is pretrained once per hospital, allowing candidate architectures to be instantiated as inherited subnetworks and evaluated through fine-tuning rather than independent pretraining, thereby improving the computational efficiency of candidate evaluation. It incorporates a two-layer cross-hospital architecture prior. The first layer retrieves high-performing source architectures using task descriptors, providing concrete examples of promising designs. The second uses SHapley Additive exPlanations (SHAP)-based meta-regression to summarize task-specific component preferences across source hospitals, providing component-level guidance for candidate design. These priors guide a multi-agent LLM search using validation feedback from the target hospital. Across six clinical prediction tasks evaluated at one held-out OneFlorida+ site and one external MIMIC-IV site, ATHENA significantly outperforms all four baselines in 9 of 12 site--task evaluations under a strict equal-compute comparison. To isolate search strategy performance, all methods use the same pretrained AutoFormer supernet for candidate evaluation; ATHENA ranks first in 9 of 12 evaluations at a maximum search budget of 30 candidate evaluations. It also shows more consistent architecture selection across repeated searches. ATHENA provides a practical approach for reducing manual architecture tuning in Transformer-based EHR modeling.

\section*{Author summary}
Electronic health records can help researchers develop models that predict patient outcomes, such as death or readmission. However, choosing a suitable model design often requires costly trial and error, and a design that works well at one hospital may not work as well at another. We developed an approach that uses lessons from earlier model searches at other hospitals to guide the design of new prediction models. We use artificial intelligence to propose and assess designs, while adapting and evaluating each candidate using local hospital data. We also reuse parts of a model initially trained once at each hospital to reduce repeated training. We tested our approach on six prediction tasks using data from two hospitals. Under the same computing budget, our approach significantly outperformed all four comparison methods in nine of twelve hospital and task combinations. Our findings suggest that knowledge about model design can be reused across hospitals to support more efficient development of prediction models.

\section{Introduction}\label{sec_intro}
The increasing availability of large-scale electronic health record (EHR) data has motivated the development of models that can learn predictive patterns directly from routinely collected clinical records \cite{wang2024recent}. Deep learning has become a standard approach for these tasks, with Transformer-based models increasingly adopted for their ability to capture dependencies among medical events across a patient's longitudinal record \cite{ren2025comprehensive}. A common two-stage paradigm consists of pretraining a Transformer backbone on large-scale EHR data to learn generalizable clinical patterns, followed by fine-tuning to adapt the model to task-specific objectives \cite{rasmy2021med,yang2023transformehr,poulain2024graph,li2026dt}.

Despite these advances, Transformer architectures for EHR modeling are still largely determined through manual design, relying on prior experience and trial-and-error. The optimal architecture can vary across prediction tasks and health systems, as differences in cohort composition, clinical workflows, and outcome characteristics may favor different architectural configurations \cite{kirchler2026large}. Consequently, a single hand-designed architecture may not perform consistently well across tasks and sites, whereas manually redesigning the architecture for each new setting is computationally costly and fails to systematically leverage knowledge gained from previous settings.

Neural architecture search (NAS) addresses this problem by treating architecture design as an optimization problem over a predefined search space~\cite{salmani2025systematic,elsken2019neural}. Classical NAS approaches, including reinforcement learning~\cite{zoph2016neural}, evolutionary algorithms~\cite{real2019regularized}, differentiable search~\cite{liu2018darts}, and Bayesian optimization~\cite{kandasamy2018neural}, have been successfully applied to vision tasks. NAS has also been applied to clinical data modeling. MUFASA~\cite{xu2021mufasa} addresses multimodal modeling through evolutionary search with a Transformer-based seed architecture, jointly optimizing modality-specific networks and fusion strategies. It also demonstrates architecture transfer between diagnosis prediction tasks. AutoDP~\cite{cui2024autodp} uses surrogate-model optimization with progressive sampling to jointly search disease-task groupings and shared encoder architectures. Evolutionary NAS for medical records~\cite{zhang2023chineseenas} optimizes bidirectional long short-term memory architectures and training hyperparameters for named-entity recognition and relation extraction, while keeping the BERT architecture fixed in the entity-recognition model. Its focus is clinical-text extraction rather than outcome prediction from longitudinal EHRs. These studies illustrate the use of NAS for multimodal fusion, multi-task prediction, and clinical-text extraction. However, systematic architecture search for pretrained Transformers in longitudinal EHR outcome prediction remains relatively underexplored.

Alongside these task-specific approaches, large language models (LLMs) have been explored as flexible NAS controllers that propose and refine architectures using task descriptions, search-space specifications, computational budgets, and evaluation feedback~\cite{ji2025rz,zheng2023can,li2026collm,su2025large}. This reduces the need to hand-design search heuristics. However, the aforementioned approaches primarily adapt proposals within the current search and do not explicitly incorporate architecture--performance evidence from previous searches at other hospitals as a reusable prior. As a result, useful architectural patterns identified in earlier hospital--task settings are not systematically incorporated into subsequent searches.

A second, complementary challenge is the computational cost of candidate evaluation. Although pretraining strategies, such as medical code-level masked language modeling~\cite{devlin2019bert}, play an important role in improving the predictive performance of Transformer-based EHR models, independently pretraining each candidate architecture on large-scale longitudinal datasets can dominate the computational cost of model development~\cite{rasmy2021med}. Weight-sharing NAS reduces this burden by training a single over-parameterized supernet whose subnetworks share parameters. Candidate architectures can then inherit the corresponding subnetwork weights and be evaluated through fine-tuning, avoiding repeated independent pretraining~\cite{guo2020single}. This makes weight-sharing evaluation particularly well suited to architecture search for pretrained EHR Transformers.

In this study, we propose ATHENA (\textbf{A}gentic \textbf{T}ransfer across \textbf{H}ospitals for \textbf{E}HR \textbf{N}eural \textbf{A}rchitecture Search), a knowledge-guided agentic NAS framework for Transformer-based EHR modeling. ATHENA addresses these two challenges by combining reusable cross-hospital architecture priors with weight-sharing candidate evaluation. We construct an AutoFormer-style supernet~\cite{chen2021autoformer} that is pretrained once per hospital and reused throughout architecture search. Each candidate inherits the corresponding subnetwork weights and is evaluated through fine-tuning. To guide candidate selection, ATHENA combines a multi-agent LLM search with a two-layer architecture prior derived from previously studied hospitals. The first layer retrieves high-performing architectures based on task descriptors, while the second summarizes architectural component effects using SHapley Additive exPlanations (SHAP)-based meta-regression. Together with target-hospital validation feedback, these priors allow the search to build on previously observed architectural patterns while retaining local model selection.

Across six clinical prediction tasks and two target health systems, ATHENA matches or outperforms conventional and LLM-guided NAS baselines under limited search budgets and produces more consistent architecture selections across repeated searches. Its weight-sharing evaluation achieves up to a 15.3-fold speedup over independently pretraining and fine-tuning candidate architectures.

\subsection{Related work}\label{sec_rel_work}

\subsubsection{Transformer-based EHR modeling}\label{sec_rel_work_1}

The longitudinal and irregular nature of EHR data makes modeling patient trajectories particularly challenging. Transformer architectures \cite{vaswani2017attention} have consequently become widely used for EHR modeling because of their ability to capture dependencies across longitudinal clinical events. BEHRT \cite{li2020behrt} represents medical codes as tokens and models longitudinal patient histories using Transformer encoders, while Med-BERT \cite{rasmy2021med} demonstrates the effectiveness of large-scale pretraining on millions of patient records.

Subsequent work has incorporated richer temporal and clinical information into Transformer-based EHR models. CEHR-BERT \cite{pang2021cehr} introduces temporal-aware embeddings and auxiliary learning objectives, whereas Hi-BEHRT \cite{li2022hi} employs a hierarchical architecture to capture long-range dependencies in extended patient histories. More recent studies have explored Transformer architectures beyond the conventional encoder-only design. TransformEHR \cite{yang2023transformehr} adopts an encoder--decoder architecture with generative pretraining objectives for longitudinal clinical modeling, while Foresight \cite{kraljevic2024foresight} employs autoregressive pretraining to model patient trajectories. CLMBR \cite{wornow2307ehrshot} learns transferable patient representations through autoregressive next-code prediction.

Collectively, these studies demonstrate the value of pretraining and contextualized representation learning for clinical prediction. Despite these advances, the architectures of Transformer-based EHR models remain largely handcrafted, with model configurations typically selected through manual tuning or limited hyperparameter search. This limitation motivates the use of NAS to systematically explore Transformer architectures for downstream clinical prediction tasks.

\subsubsection{NAS for healthcare models}\label{sec_rel_work_2}

NAS automates architecture design by defining a search space, a search strategy, and a performance-estimation scheme \cite{white2023neural}. Recent advances, including gradient-based optimization \cite{liu2018darts} and weight-sharing supernet methods \cite{guo2020single}, have substantially reduced the computational cost of architecture search. NAS has also been extended to Transformer architectures through methods such as Evolved Transformer \cite{so2019evolved}, HAT \cite{wang2020hat}, NAS-BERT \cite{xu2021bert}, AutoBERT-Zero \cite{gao2022autobert}, and Primer \cite{so2022primer}. However, these methods were primarily developed for natural-language and vision tasks and do not explicitly account for the characteristics of longitudinal EHR data.

Within healthcare, NAS has been applied to medical imaging~\cite{su2025large,bal2025metallmix}, multimodal EHR fusion~\cite{xu2021mufasa,cui2024automated}, multi-task disease prediction~\cite{cui2024autodp}, and clinical-text information extraction~\cite{zhang2023chineseenas}. MUFASA~\cite{xu2021mufasa} jointly searches modality-specific networks and fusion strategies using a Transformer-based seed, and evaluates architecture transfer between diagnosis tasks. AutoFM~\cite{cui2024automated} also searches modality encoders and fusion strategies, whereas AutoDP~\cite{cui2024autodp} jointly optimizes disease-task grouping and shared encoder architectures. The evolutionary method for medical records searches bidirectional long short-term memory architectures and training settings for entity and relation extraction, with a fixed BERT architecture~\cite{devlin2019bert} in the entity-recognition model. These studies address complementary design problems. ATHENA focuses on reusing evaluated source-hospital architectures and component--performance associations to guide local selection of pretrained Transformer configurations for longitudinal clinical outcome prediction.

\subsubsection{LLM-driven NAS}\label{sec_rel_work_3}

Recent work has explored the use of LLMs for NAS, leveraging architectural knowledge encoded during pretraining to generate and refine candidate architectures. Early studies demonstrated that LLMs can serve as architecture generators or search controllers within NAS pipelines. GENIUS \cite{zheng2023can} employs an LLM as a black-box architecture optimizer that iteratively proposes and refines candidate architectures through natural-language interaction, while EvoPrompting \cite{chen2023evoprompting} integrates LLM-based mutation and crossover operators into an evolutionary search framework. Subsequent work has further combined LLM reasoning with conventional search strategies. GPT-NAS \cite{yu2023gpt} couples GPT-guided architecture generation with evolutionary optimization, whereas LLMatic \cite{nasir2024llmatic} integrates LLM-based code generation with quality-diversity search to discover diverse and robust neural architectures.

More recent approaches explicitly use architectural knowledge and search history. RZ-NAS~\cite{ji2025rz} incorporates reflection and zero-cost evaluation, while CoLLM-NAS~\cite{li2026collm} coordinates multiple LLM agents using architectural priors, iterative feedback, and historical trajectories. In healthcare, Pathology-NAS~\cite{su2025large} combines LLM-guided search with a pretrained supernet to discover lightweight architectures across medical imaging tasks. These methods demonstrate the value of priors, feedback, and weight-sharing evaluation. ATHENA investigates a complementary source of guidance: architecture--performance evidence accumulated across hospitals and clinical prediction tasks. It integrates retrieved source architectures and architecture-effect summaries with target-hospital validation to guide local selection of pretrained longitudinal EHR Transformer configurations.

\section{Materials and methods}\label{sec_methods}

\subsection{Ethics statement}
The University of Florida Institutional Review Board approved this study (IRB202400190), which was non-human subject research involving only the collection of existing, de-identified patient medical data. The requirement for informed consent was waived.

\subsection{Overview}\label{sec_methods_1}

\begin{figure}[H]
\centering
\includegraphics[width=\linewidth,height=0.72\textheight,keepaspectratio]{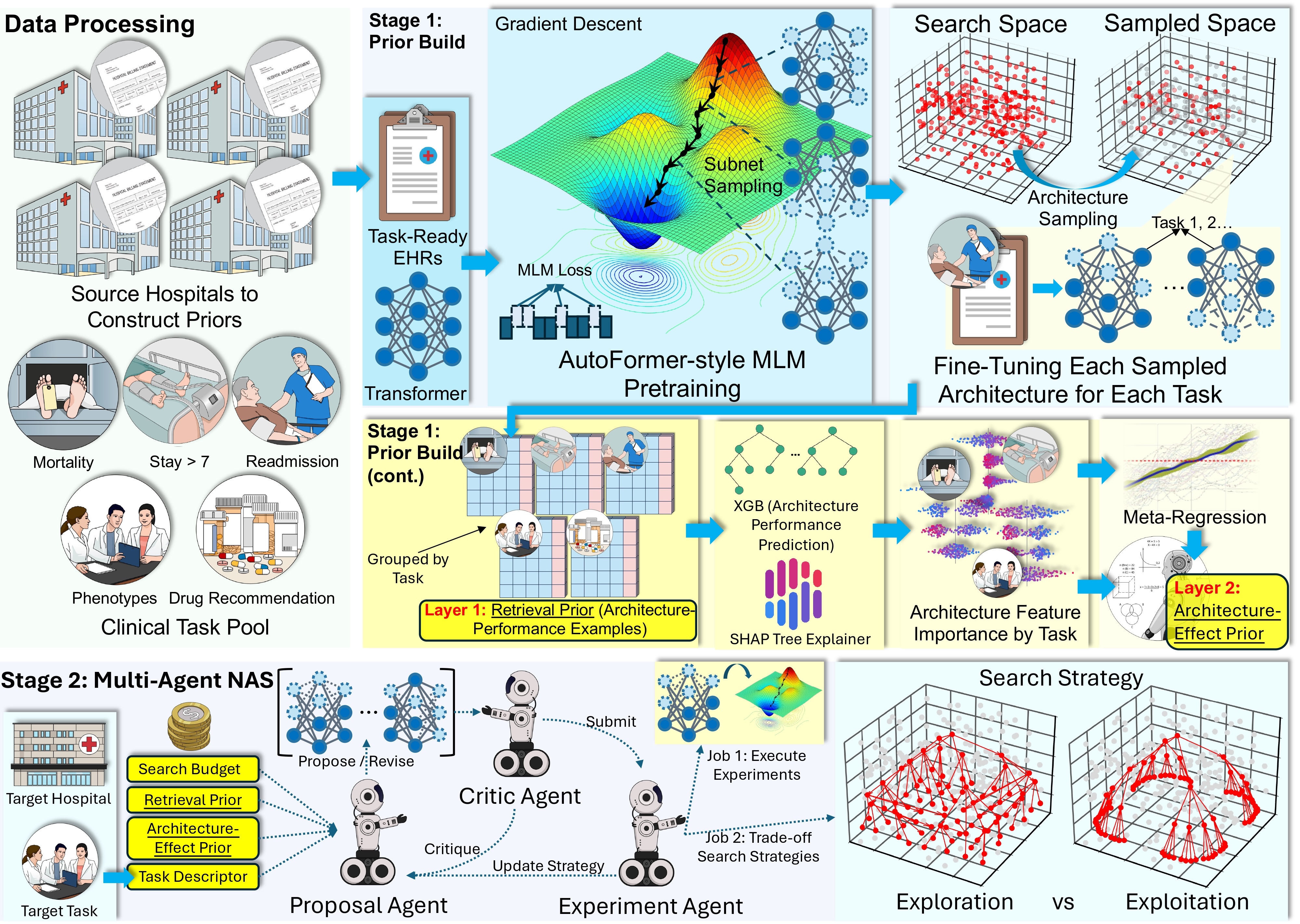}
\caption{The ATHENA framework. \emph{Stage 1} evaluates sampled architectures across source hospitals and tasks to obtain architecture--performance metadata. The metadata yield a retrieval prior of high-performing source architectures and a SHAP-based architecture-effect prior. \emph{Stage 2} combines these priors with target-validation feedback to guide agentic NAS while adaptively balancing \emph{exploration} and \emph{exploitation}.}
\label{fig:workflow}
\end{figure}

ATHENA is a two-stage framework for knowledge-guided NAS of EHR Transformers (Fig~\ref{fig:workflow}). Let $\mathcal{S}$ denote the source hospitals, $H$ a held-out target hospital, and $\mathcal{T}$ the clinical prediction tasks.

In \textit{Stage 1}, ATHENA pretrains one hospital-specific AutoFormer-style supernet at each $S\in\mathcal{S}$ using masked language modeling (MLM)~\cite{rasmy2021med} and evaluates a fixed collection of subnet architectures on the source tasks. Each metadata record has the form
\begin{equation}
\label{eq:source-metadata-record}
\left(S,t,a,\operatorname{Params}_{S,t}(a),\mathbf{m}^{\mathrm{src}}_{S,t}(a)\right),
\end{equation}
where $t\in\mathcal{T}$, $a$ is an architecture, $\operatorname{Params}_{S,t}(a)$ is its parameter count, and $\mathbf{m}^{\mathrm{src}}_{S,t}(a)$ contains its source performance metrics. Pooled source records yield two complementary priors: \emph{Layer 1} retrieves concrete high-performing architectures from a task-matched source hospital, whereas \emph{Layer 2} summarizes cross-hospital associations between architectural choices and performance.

In \textit{Stage 2}, a \emph{Proposal Agent} generates candidates, a \emph{Critic Agent} checks and refines them, and an \emph{Experiment Agent} evaluates accepted subnetworks and chooses whether the next round should emphasize \emph{exploration} or \emph{exploitation}. Both prior layers provide soft guidance together with feedback from earlier target-hospital evaluations.

\subsection{Problem formulation}\label{sec_methods_2}\label{sec_methods_3}

Let $\mathcal{C}$ denote the vocabulary of diagnosis, medication, laboratory, and procedure tokens. The chronologically ordered visits of patient $p$ are
\begin{equation}
\label{eq:patient-visits}
\mathcal{V}_p=(v_{p,1},\ldots,v_{p,T_p}), \qquad
v_{p,\tau}=\{c_{p,\tau,1},\ldots,c_{p,\tau,N_{p,\tau}}\},
\end{equation}
where $c_{p,\tau,i}\in\mathcal{C}$. Visits are chronological, but no within-visit order is assumed because event timestamps within an encounter may not reflect the underlying clinical sequence \cite{choi2016multi,rasmy2021med,zhang2020diagnostic}. For each prediction example, only visits available by its index time are retained. They are flattened into a token sequence, prepended with \mbox{\texttt{[CLS]}}, and annotated with token-type and visit-index embeddings.

For target hospital $H$ and task $t$, a candidate Transformer architecture is defined as:
\begin{equation}
\label{eq:architecture-tuple}
a=(d,L,\gamma,h),
\end{equation}
where $d$ is the embedding dimension, $L$ is the number of encoder layers, $\gamma$ is the multi-layer perceptron (MLP) expansion ratio, and $h$ is the number of attention heads. Let $\mathcal{D}$, $\mathcal{L}$, $\mathcal{R}$, and $\mathcal{H}$ denote their respective candidate sets. The unfiltered search space is
\begin{equation}
\label{eq:architecture-search-space}
\mathcal{A}_0
={}
\mathcal{D}\times\mathcal{L}\times\mathcal{R}\times\mathcal{H},
\end{equation}
in the order $(d,L,\gamma,h)$. The target-specific legal set is
\begin{equation}
\label{eq:legal-search-space}
\mathcal{A}_{H,t}
={}
\left\{
\begin{aligned}
a\in\mathcal{A}_0:\;&d\bmod h=0,\\
&\operatorname{Params}_{H,t}(a)\leq P_{\max}
\end{aligned}
\right\},
\end{equation}
where $P_{\max}$ is the parameter-count limit. The divisibility condition ensures an integral attention-head dimension.

During target search, each evaluated architecture produces a validation metric vector comprising Accuracy, $F_1$, the area under the receiver operating characteristic curve (AUROC), and the area under the precision--recall curve (AUPRC):
\begin{equation}
\label{eq:validation-metric-set}
\mathcal{G}=\{\mathrm{Accuracy},F_1,\mathrm{AUROC},\mathrm{AUPRC}\}.
\end{equation}
For binary tasks, $F_1$ is the positive-class binary $F_1$, and AUROC and AUPRC use the positive-class probability. For multilabel tasks, all metrics are macro-averaged over label classes. After $b$ accepted evaluations, let
\begin{equation}
\label{eq:validation-history}
\mathcal{E}_b
=
\left(\left(a_i,\mathbf{m}_{\mathrm{val},t}(a_i)\right)\right)_{i=1}^{b}
\end{equation}
denote the ordered target-validation history. ATHENA ranks each metric in descending performance order, assigns average ranks to ties, and computes
\begin{align}
R_t(a;\mathcal{E}_b)
&=
\frac{1}{|\mathcal{G}|}
\sum_{g\in\mathcal{G}}
\operatorname{rank}^{\downarrow}_{g,t}(a;\mathcal{E}_b),
\label{eq:athena-composite-rank}
\end{align}
where lower ranks are preferred. If composite ranks tie, the architecture evaluated first is retained, making selection deterministic with respect to the archive order.

Let $B$ denote the maximum evaluation budget, and let $b_{\mathrm{end}}\leq B$ denote the number of architectures evaluated before budget exhaustion or early termination. ATHENA returns the best-performing evaluated architecture:
\begin{equation}
\label{eq:selected-architecture}
a^\star_{H,t}
=
\argmin_{a:\,(a,\mathbf{m})\in\mathcal{E}_{b_{\mathrm{end}}}}
R_t(a;\mathcal{E}_{b_{\mathrm{end}}}).
\end{equation}

\subsection{Transformer subnet architecture}\label{sec_methods_4}

Given architecture $a=(d,L,\gamma,h)$, token $i$ is represented by the sum of its code, token-type, and visit-index embeddings:
\begin{equation}
\label{eq:token-embedding}
\mathbf{h}^{0}_i
=
\mathbf{e}^{\mathrm{code}}_i
+
\mathbf{e}^{\mathrm{type}}_i
+
\mathbf{e}^{\mathrm{visit}}_i,
\end{equation}
where $\mathbf{e}^{\mathrm{code}}_i$, $\mathbf{e}^{\mathrm{type}}_i$, and $\mathbf{e}^{\mathrm{visit}}_i$ denote the corresponding embedding vectors. 

The input sequence consists of a prepended \mbox{\texttt{[CLS]}} token followed by the flattened EHR token sequence:
\begin{equation}
\label{eq:input-representation}
\mathbf{H}^{0}
=
\left[
\mathbf{h}^{0}_{\mbox{\texttt{[CLS]}}},
\mathbf{h}^{0}_{1},
\ldots,
\mathbf{h}^{0}_{n}
\right].
\end{equation}

ATHENA uses a pre-normalization Transformer encoder. Multi-head attention (MHA), feed-forward network (FFN), and layer normalization (LN) denote the corresponding Transformer operations below. For layer $\ell=1,\ldots,L$,
\begin{align}
\widetilde{\mathbf{H}}^{\ell}
&=
\mathbf{H}^{\ell-1}
+
\operatorname{MHA}_{a,\ell}
\!\left(\operatorname{LN}^{\mathrm{attn}}_{a,\ell}(\mathbf{H}^{\ell-1})\right),
\label{eq:attention-residual-update}\\
\mathbf{H}^{\ell}
&=
\widetilde{\mathbf{H}}^{\ell}
+
\operatorname{FFN}_{a,\ell}
\!\left(\operatorname{LN}^{\mathrm{ffn}}_{a,\ell}(\widetilde{\mathbf{H}}^{\ell})\right),
\label{eq:ffn-residual-update}
\end{align}
where $\operatorname{MHA}_{a,\ell}$ uses $h$ heads and $\operatorname{FFN}_{a,\ell}$ has hidden width $\gamma d$. Dropout and drop-path are applied within the residual branches. 

For downstream prediction, the hidden representation corresponding to the final \mbox{\texttt{[CLS]}} token is used as the patient-level representation:
\begin{equation}
\label{eq:patient-representation}
\mathbf{z}_{p,j,a}
=
\LN_{a}\!\left(\mathbf{h}^{L}_{p,j,\mathrm{[CLS]}}\right),
\end{equation}
where $\mathbf{h}^{L}_{p,j,\mathrm{[CLS]}}$ is the final \mbox{\texttt{[CLS]}} hidden state for target admission $j$.

\subsection{AutoFormer-style supernet training and subnet evaluation}\label{sec_methods_5}

\begin{figure}[H]
\centering
\includegraphics[width=\linewidth,height=0.72\textheight,keepaspectratio]{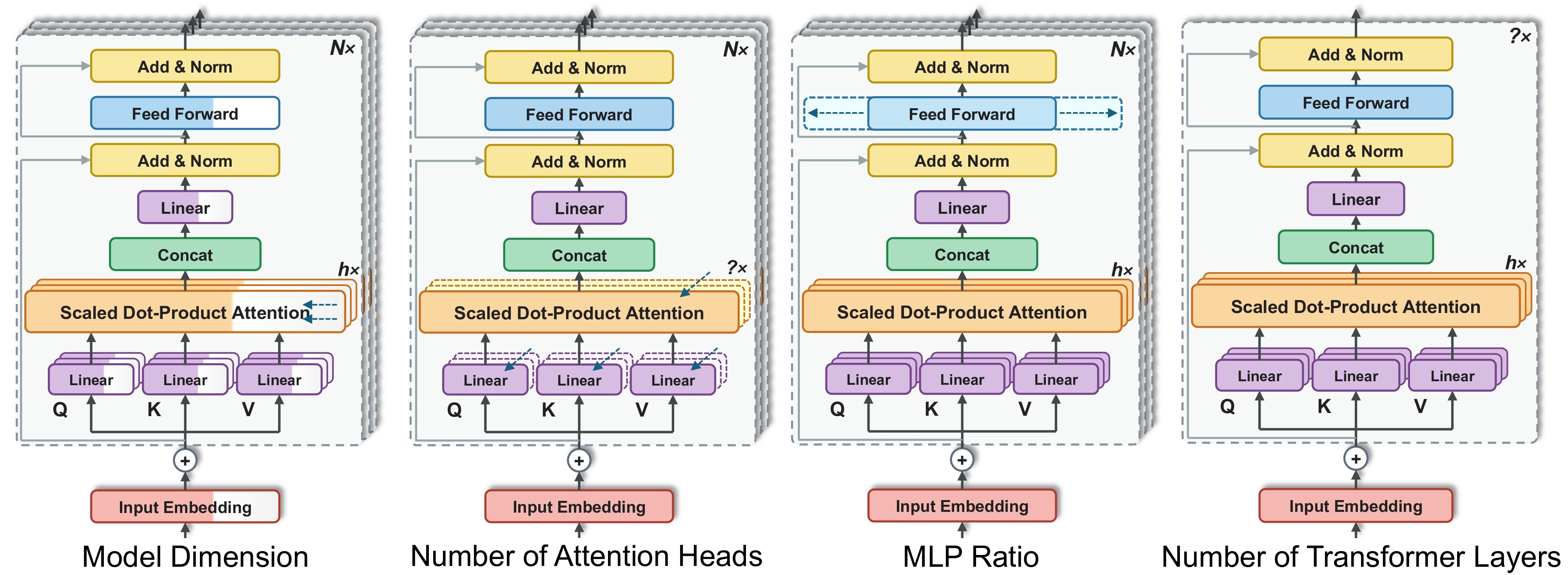}
\caption{Illustration of the four configurable Transformer architecture parameters in the NAS space.}
\label{fig:feature_space}
\end{figure}

Training every architecture independently would require repeated self-supervised pretraining and downstream fine-tuning. ATHENA therefore uses an AutoFormer-style shared-weight supernet~\cite{chen2021autoformer} spanning the search space $\mathcal{A}_0$. The supernet takes the componentwise maximal configuration in the search space. A subnet activates the first $d$ embedding channels, the first $L$ encoder blocks, and the first $\gamma d$ FFN units. Its attention module partitions fixed internal query, key, and value representations into $h$ heads, while the input and output projections are sliced to width $d$. Thus, each candidate architecture is instantiated as a weight-inherited subnetwork of the shared supernet rather than as an independently trained model. Fig~\ref{fig:feature_space} illustrates how the four configurable parameters alter the resulting subnet. 

In the shared-supernet comparison, one supernet is pretrained per hospital and shared by all methods at that hospital. Each MLM minibatch samples $a\sim\operatorname{Uniform}(\mathcal{A}_0)$, activates the corresponding parameter slices, and updates only those active weights. During downstream fine-tuning, $a$ is fixed and the same inherited slices are updated at every epoch.

Following BERT-style masking~\cite{devlin2019bert}, a subset of non-special tokens is selected and corrupted for MLM. Let $\Omega$ denote the set of masked positions. The pretraining loss is
\begin{equation}
\label{eq:mlm-loss}
\mathcal{L}_{\mathrm{MLM}}
=
-
\sum_{i\in\Omega}
\log
p_{\mathbf{W}_{a}}
\!\left(
c_i \mid \widetilde{\mathbf{H}}^0
\right),
\end{equation}
where $\widetilde{\mathbf{H}}^0$ is constructed from the corrupted tokens and $\mathbf{W}_{a}$ denotes the active subnet weights. Selected non-special tokens are replaced by \mbox{\texttt{[MASK]}}, replaced by a random token of the same modality, or left unchanged.

Each candidate inherits its encoder weights from this checkpoint, is fine-tuned on the target training split, and is monitored on the target validation split.

\subsection{Cross-hospital prior construction}\label{sec_methods_6}

ATHENA constructs two complementary source-derived priors (Table~\ref{tab:prior-layers}). \emph{Layer 1} retrieves top-ranked observed configurations, providing concrete examples of high-performing architectures; \emph{Layer 2} summarizes component preferences across source hospitals, providing guidance for individual architectural choices. Both priors provide soft guidance, while final architecture selection depends on target-validation performance.

\begin{table}[H]
\singlespacing
\small
\centering
\setlength{\tabcolsep}{3pt}
\renewcommand{\arraystretch}{1.15}
\caption{Complementary roles of the two cross-hospital prior layers.}
\label{tab:prior-layers}
\begin{tabular}{@{}>{\raggedright\arraybackslash}p{0.11\linewidth}>{\raggedright\arraybackslash}p{0.42\linewidth}>{\raggedright\arraybackslash}p{0.42\linewidth}@{}}
\toprule
Aspect & \emph{Layer 1}: task-driven retrieval & \emph{Layer 2}: architecture-effect prior \\
\midrule
Input & Source and target training-set task descriptors; evaluated source architecture--performance records. & Architecture--performance records for the same task across source hospitals. \\
Goal & Retrieve relevant, high-performing architecture examples. & Estimate cross-hospital performance associations for architectural choices. \\
Output & Top $K=5$ evaluated source configurations and their metrics. & SHAP-based level preferences, confidence intervals, feature importance, and pairwise rules. \\
Use & Concrete examples for candidate proposal and refinement. & Directional evidence for candidate proposal and refinement. \\
\bottomrule
\end{tabular}
\end{table}

For example, for mortality prediction, \emph{Layer 1} compares the target task descriptor with source mortality descriptors and retrieves five high-ranked, previously evaluated configurations from the most similar source hospital. \emph{Layer 2} instead pools mortality records across the four source hospitals to summarize associations between architectural choices and performance. The downstream agents (Section~\ref{sec_methods_7}) use both the examples and component-level guidance to propose valid candidate architectures, whose suitability is assessed on the target validation set. 

\subsubsection{Task-driven retrieval prior}

For task $t$ at hospital $H$, the task descriptor is
\begin{equation}
\label{eq:task-descriptor}
\mathbf{g}_{H,t}=
\left[
\mathbb{I}_{\mathrm{binary}}(t),
\mathbb{I}_{\mathrm{multilabel}}(t),
\widetilde C_t,
\mathcal{H}_{\mathrm{label}}(H,t),
\pi^+_{H,t},
\widetilde\Delta_t
\right],
\end{equation}
where the first two entries indicate task type, $\widetilde C_t=\log(C_t)/\log(20)$ is the normalized class count, $\mathcal{H}_{\mathrm{label}}$ is the label entropy, $\pi^+$ is the positive-label prevalence, and $\widetilde\Delta_t=\Delta_t/365$ is the normalized prediction horizon. For multilabel tasks, $\mathcal{H}_{\mathrm{label}}$ and $\pi^+$ are averaged across classes.

Label entropy and prevalence in the task descriptor are computed on the training split only, for both source and target hospitals; validation and test labels are never used for descriptor construction.

When exact task $t$ is present at the source hospitals, ATHENA selects
\begin{equation}
\label{eq:source-hospital-selection}
S_t^\star
=
\argmax_{S\in\mathcal{S}}
\frac{
\mathbf{g}_{H,t}^{\mathsf{T}}\mathbf{g}_{S,t}
}{
\lVert\mathbf{g}_{H,t}\rVert_2
\lVert\mathbf{g}_{S,t}\rVert_2
}.
\end{equation}
Here $\lVert\cdot\rVert_2$ denotes the Euclidean norm. ATHENA then ranks the legal source architectures for $(S_t^\star,t)$ by the composite source-performance rank corresponding to \Eqref{eq:athena-composite-rank} and retrieves the top $K$. These configurations and their source metrics form the \emph{Layer 1} retrieval set $\mathcal{A}_{\mathrm{ret}}$, with $K=5$ in our experiments. The ranking is limited to previously evaluated source architectures and therefore does not guarantee the optimal architecture in the full source search space.

If exact task metadata are unavailable, ATHENA selects the source hospital with the highest cosine similarity between standardized dataset profiles (e.g., sample size, modality counts, and mean encounters per patient). Within the selected hospital, ATHENA chooses the available surrogate task whose descriptor is most similar to $\mathbf{g}_{H,t}$.

\subsubsection{Architecture-effect prior}

\emph{Layer 2} identifies architectural component levels and combinations associated with better task-specific performance across source hospitals. Architecture evaluations are conducted independently within each source hospital, rather than through joint cross-hospital optimization. For task $t$, evaluation records from all source hospitals are pooled, and the performance target for record $i$ is defined as
\begin{equation}
\label{eq:source-performance-target}
y_{t,i}
=
-
R_t\!\left(a_i;\mathcal{E}_{\mathcal{S},t}\right),
\end{equation}
where $\mathcal{E}_{\mathcal{S},t}$ denotes the pooled source metadata table for task $t$, and larger $y_{t,i}$ values indicate better relative performance among source records. Hospital identity is retained for subsequent mixed-effects analysis.

For each task, ATHENA fits an XGBoost surrogate~\cite{chen2016xgboost}:
\begin{equation}
\label{eq:xgboost-surrogate}
y_{t,i}
=
f_{\mathrm{XGBoost},t}(a_i)
+
\varepsilon_{t,i}.
\end{equation}
Discrete architectural features are supplied to the tree model in numeric form. SHAP TreeExplainer~\cite{lundberg2020local2global} then computes the signed contribution $\phi^{(q)}_{t,i}$ of each architectural feature $q$ for record $i$.

To distinguish cross-hospital architectural patterns from hospital-specific variation, ATHENA fits a mixed-effects model:
\begin{equation}
\label{eq:shap-mixed-effects}
\phi^{(q)}_{t,i}
=
\mu^{(q)}_{t,\ell_{i,q}}
+
u^{(q)}_{t,S_i}
+
\epsilon^{(q)}_{t,i},
\qquad
u^{(q)}_{t,S}
\sim
\mathcal{N}(0,\sigma_{u,q}^2),
\end{equation}
where $\ell_{i,q}$ denotes the categorical level of feature $q$, and $u^{(q)}_{t,S_i}$ is a hospital-specific random intercept. The level-specific mean SHAP contributions $\mu^{(q)}_{t,\ell}$ are reconstructed from the reference-coded model.

Component levels whose confidence intervals lie entirely above or below zero are classified as \emph{preferred} or \emph{discouraged}, respectively; the remaining levels are classified as \emph{inconclusive}. The same mixed-effects analysis is applied to the most influential feature pair to identify favorable and unfavorable component combinations. Feature importance, level classifications, confidence intervals, and component combination rules collectively constitute the \emph{Layer 2} prior $\mathcal{P}_{\mathrm{meta}}$, providing agents with non-causal, directional guidance for architecture search.

\subsection{Agentic NAS}\label{sec_methods_7}

ATHENA coordinates a \emph{Proposal Agent} $A_p$, a \emph{Critic Agent} $A_c$, and an \emph{Experiment Agent} (Table~\ref{tab:agent-roles}). The \emph{Experiment Agent} consists of a deterministic evaluator $A_e$ and an LLM-based strategy module $A_s$. Proposal, critique, revision, and strategy selection are LLM calls; legality checks, subnet fine-tuning, metric calculation, ranking, and final selection are controller operations.

\begin{table}[H]
\singlespacing
\footnotesize
\centering
\setlength{\tabcolsep}{3pt}
\renewcommand{\arraystretch}{1.18}
\caption{Roles of the three agents in ATHENA.}
\label{tab:agent-roles}
\begin{tabular*}{\linewidth}{@{\extracolsep{\fill}}>{\raggedright\arraybackslash}p{0.95in}>{\raggedright\arraybackslash}p{1.05in}>{\raggedright\arraybackslash}p{2.70in}>{\raggedright\arraybackslash}p{1.38in}@{}}
\toprule
\textbf{Agent} & \textbf{Implementation} & \textbf{Main function} & \textbf{Output} \\
\midrule
\emph{Proposal Agent} $A_p$
& LLM
& Proposes architectures using task information, cross-hospital priors, search history, and the current strategy; revises proposals based on critique feedback.
& Candidate architectures and design rationales. \\
\addlinespace[3pt]
\emph{Critic Agent} $A_c$
& LLM
& Reviews proposed designs for constraint compatibility and novelty, flags concerns about prior preferences, and recommends revisions.
& Structured critiques and revision recommendations. \\
\addlinespace[3pt]
\emph{Experiment Agent}
& Deterministic evaluator $A_e$ and LLM strategy module $A_s$
& $A_e$ fine-tunes accepted subnetworks and computes target-validation metrics; $A_s$ uses updated search feedback to select \emph{exploration} or \emph{exploitation} for the next round.
& Validation records and the next-round strategy with its rationale. \\
\bottomrule
\end{tabular*}
\par\smallskip
{\raggedright\footnotesize LLM, large language model. Proposal, critique, revision, and strategy selection use LLM calls. Legality checks, subnet fine-tuning, metric calculation, candidate ranking, and final architecture selection are deterministic controller operations.\par}
\end{table}

For each target hospital-task pair $(H,t)$, ATHENA constructs a fixed context
\begin{equation}
\label{eq:agent-search-context}
\mathcal{C}_{H,t}
= 
\{\mathbf{g}_{H,t},\mathcal{A}_{\mathrm{ret}},\mathcal{P}_{\mathrm{meta}}\},
\end{equation}
containing the target task descriptor, the $K$ retrieved architectures, and the \emph{Layer 2} prior.

At round $r$, ATHENA maintains two distinct histories. The \emph{ordered search memory} (or search transcript) $\mathcal{M}^{(r)}$ contains proposals, critiques, revisions, evaluation feedback, and strategy decisions, and is supplied as LLM context. The \emph{structured validation archive}
\begin{equation}
\label{eq:round-validation-archive}
\mathcal{E}^{(r)}
=
\left(\left(a_i,\mathbf{m}_{\mathrm{val},t}(a_i)\right)\right)_{i=1}^{b_r}
\end{equation}
contains only evaluated architectures and their validation metrics. It is used by the deterministic controller for ranking and selection. Thus, $\mathcal{M}^{(r)}$ carries accumulated search experience, whereas $\mathcal{E}^{(r)}$ is the authoritative record of empirical evidence.

The round also has remaining budget $B-b_r$ and a strategy
\begin{equation}
\label{eq:search-strategy}
s^{(r)}=(z^{(r)},\eta^{(r)}),
\qquad
z^{(r)}\in\{\textit{exploration},\textit{exploitation}\},
\end{equation}
where $\eta^{(r)}$ is a natural-language rationale. \emph{Exploration} encourages coverage of under-sampled architectural choices, whereas \emph{exploitation} focuses proposals around configurations that have performed well on target validation.

\begin{figure}[H]
\centering
\includegraphics[width=\linewidth,height=0.72\textheight,keepaspectratio]{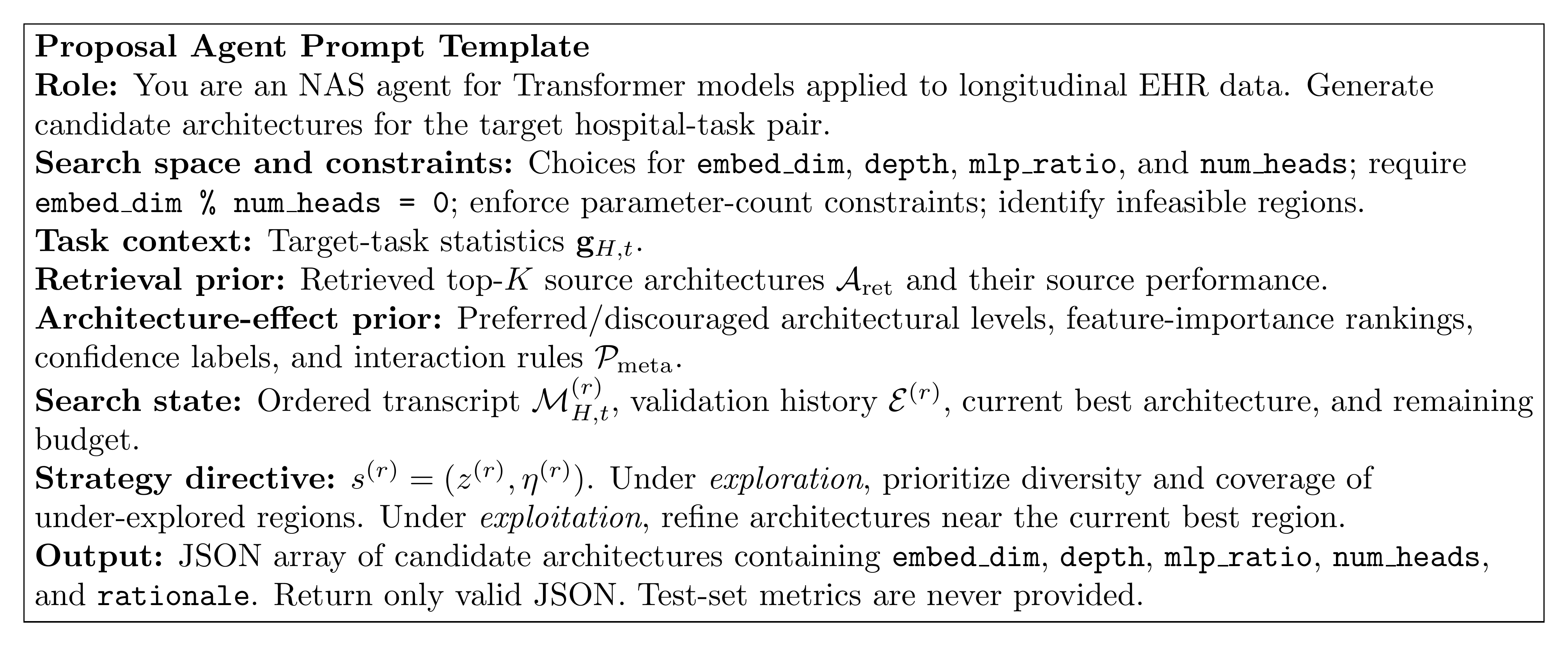}
\caption{The \emph{Proposal Agent} prompt template. The implementation populates each block with task-specific values before invoking the LLM.}
\label{fig:athena-prompt-template}
\end{figure}

Conditioned on $(\mathcal{C}_{H,t},\mathcal{M}^{(r)},\mathcal{E}^{(r)},B-b_r,s^{(r)})$, $A_p$ proposes candidates with rationales. The \emph{Critic} checks legality and novelty, provides structured feedback, and allows rejected non-duplicate candidates to be revised for at most $R$ passes. Parameter-limit violations and duplicates are hard rejections; disagreement with a \emph{Layer 2} preference is only a soft concern. The proposal--critique--revision records are appended to $\mathcal{M}^{(r)}$ even when no candidate is evaluated. Fig~\ref{fig:athena-prompt-template} shows the \emph{Proposal Agent} prompt template.

The evaluator fine-tunes each accepted subnet and appends the resulting records to both $\mathcal{E}^{(r)}$ and $\mathcal{M}^{(r)}$. The controller then recomputes the leader using \Eqref{eq:athena-composite-rank}. When budget remains, $A_s$ reads the updated transcript and validation trajectory and selects the next strategy, treating the cross-hospital prior as secondary evidence. The strategy decision is also appended to $\mathcal{M}^{(r)}$.

If a round produces no legal, non-duplicate candidate, no evaluation budget is consumed and a consecutive-failure counter is incremented. The counter resets after any successful evaluation round. Search terminates when the budget $B$ is exhausted or after $F_{\max}$ consecutive empty rounds, and ATHENA returns the current validation-selected leader. A run in which no candidate is evaluated is recorded as unsuccessful. Algorithm~\ref{alg:athena-search} summarizes this search logic.
\begin{algorithm}[H]
\caption{ATHENA agentic NAS loop}
\label{alg:athena-search}
\small
\begin{algorithmic}[1]

\Require Fixed context $\mathcal{C}_{H,t}$, legal set $\mathcal{A}_{H,t}$,
budget $B$, critique-pass limit $R$, failure limit $F_{\max}$
\Ensure Validation-selected architecture $a^\star_{H,t}$, or failure if no candidate is evaluated

\State Initialize ordered search memory $\mathcal{M}\leftarrow()$,
validation archive $\mathcal{E}\leftarrow()$, strategy $s\leftarrow\textit{exploration}$,
and failure count $f\leftarrow0$
\While{$|\mathcal{E}|<B$ \textbf{and} $f<F_{\max}$}
    \State $\mathcal{P}\leftarrow A_p(\mathcal{C}_{H,t},\mathcal{M},\mathcal{E},B-|\mathcal{E}|,s)$
    \State $(\mathcal{U},\Delta\mathcal{M})\leftarrow
    \operatorname{CritiqueAndRevise}(\mathcal{P},\mathcal{C}_{H,t},\mathcal{M},\mathcal{E},s,R)$
    \State $\mathcal{M}\leftarrow\mathcal{M}\mathbin{\Vert}\Delta\mathcal{M}$
    \State Remove illegal and duplicate candidates from $\mathcal{U}$ in proposal order
    \State Retain at most the first $B-|\mathcal{E}|$ candidates in $\mathcal{U}$
    \If{$\mathcal{U}=\emptyset$}
        \State $f\leftarrow f+1$; \textbf{continue}
    \EndIf
    \State $\mathcal{V}\leftarrow A_e(\mathcal{U})$ using target training and validation data
    \State $\mathcal{E}\leftarrow\mathcal{E}\mathbin{\Vert}\mathcal{V}$;
    $\mathcal{M}\leftarrow\mathcal{M}\mathbin{\Vert}\mathcal{V}$; $f\leftarrow0$
    \State $a_{\mathrm{best}}\leftarrow\argmin_{a:\,(a,\mathbf{m})\in\mathcal{E}}R_t(a;\mathcal{E})$
    \If{$|\mathcal{E}|<B$}
        \State $s\leftarrow A_s(\mathcal{C}_{H,t},\mathcal{M},\mathcal{E},
        a_{\mathrm{best}},B-|\mathcal{E}|)$
        \State $\mathcal{M}\leftarrow\mathcal{M}\mathbin{\Vert}s$
    \EndIf
\EndWhile
\If{$\mathcal{E}=\emptyset$}
    \State \Return failure
\EndIf
\State \Return $a^\star_{H,t}\leftarrow a_{\mathrm{best}}$

\end{algorithmic}
\end{algorithm}

\subsection{Experiment configuration}\label{sec_config}

\subsubsection{Data sources and cohort construction}\label{sec_config_1}
This study uses EHR data from OneFlorida+ \cite{fleurence2014pcornet,hogan2022oneflorida} and MIMIC-IV \cite{johnson2023mimic}. OneFlorida+ is a large clinical research network within the National Patient-Centered Clinical Research Network (PCORnet) \cite{fleurence2014pcornet} and contains standardized EHR data across multiple health systems. MIMIC-IV is a public de-identified EHR database from Beth Israel Deaconess Medical Center containing more than 524,000 hospital admissions from over 257,000 patients. Both datasets include longitudinal diagnoses, procedures, medications, and laboratory records used to construct the clinical prediction tasks.

For ATHENA, the processed OneFlorida+ data comprise four prior-source sites (Sites A--D) and one internal held-out target (Site E); MIMIC-IV serves as the external held-out target. Records from both target cohorts are excluded from the cross-hospital prior. The OneFlorida+ and MIMIC-IV cohorts follow the same EHR tokenization, task-construction, and split conventions.

Each hospital's unlabeled EHR corpus is first divided into a pretraining pool of patients without downstream labels and a downstream pool containing the remaining patients. The pretraining pool is further split into 90\% for masked-language-model supernet pretraining and 10\% for validation.

We evaluate six downstream clinical prediction tasks, each defined relative to the target hospital admission:

\begin{itemize}
    \item \textbf{Mortality (binary):} Predicts in-hospital death during the target admission.

    \item \textbf{Stay $>$ 7d (binary):} Predicts whether the target admission exceeds seven days.

    \item \textbf{Readmission (3M; binary):} Predicts hospital readmission within 90 days after discharge from the target admission.

    \item \textbf{Phenotype (6M; multilabel):} Predicts 18 predefined diagnosis categories at the next admission occurring within 180 days.

    \item \textbf{Phenotype (12M; multilabel):} Predicts the same 18 diagnosis categories at the next admission occurring within 365 days.

    \item \textbf{Drug Recommendation (multilabel):} Predicts 55 Anatomical Therapeutic Chemical (ATC) level-4 medication classes prescribed during the target admission, using admission diagnoses and prior visit history.
\end{itemize}

The phenotype tasks use 18 commonly benchmarked categories~\cite{xu2023hypergraph,yao2024drfuse} and include only patients with a qualifying subsequent admission. Drug Recommendation uses a separate eligibility cohort restricted to patients with at least one in-vocabulary medication. Detailed task definitions, cohort inclusion and exclusion criteria, and outcome specifications are provided in Method~\ref{supp:method-cohort} and \ref{tab:supp-outcome-defs}~Table in S1 Appendix.

Within the downstream pool, patient-level training, validation, and test splits are defined by task family: 20\%, 40\%, and 40\%, respectively, for the three binary tasks (Mortality, Stay $>$ 7d, and Readmission (3M)) and 40\%, 30\%, and 30\%, respectively, for Phenotype (6M) and Phenotype (12M). Drug Recommendation is independently split into training (40\%), validation (30\%), and test (30\%) sets within its eligibility cohort. For all tasks, the training set is used for subnet fine-tuning, the validation set guides architecture search and model selection, and the test set is evaluated only once after the search terminates.

\subsubsection{Transformer architecture search space}\label{sec_config_2}

The NAS search space comprises four configurable Transformer architecture parameters: embedding dimension, depth, number of attention heads, and MLP expansion ratio (Fig~\ref{fig:feature_space}). Embedding dimension takes values in $\{32,64,128,256\}$, while depth, number of attention heads, and MLP expansion ratio each take values in $\{1,2,4,8\}$. The same search space is used across all hospitals, tasks, and NAS methods to ensure a consistent comparison. Other implementation details are provided in Method~\ref{supp:method-implementation} in S1 Appendix.

\subsubsection{Compared methods and ablations}\label{sec_config_3}

We compare ATHENA against representative baselines from both classical and LLM-based NAS:

\begin{itemize}
    \item \textbf{Random Search.} Uniformly samples legal, non-duplicate architectures from $\mathcal{A}_{H,t}$ until the evaluation budget is exhausted.

   \item \textbf{Evolutionary algorithm (EA)}. Following regularized evolution for NAS \cite{real2019regularized}, this baseline maintains a population of candidate architectures and iteratively selects high-performing parents through tournament selection, generates new candidates via mutation, and removes the oldest individuals from the population.

    \item \textbf{GENIUS}~\cite{zheng2023can}. A single-agent LLM-based NAS framework that iteratively proposes candidate architectures based on search history and validation feedback, without the multi-agent collaboration used in ATHENA.

    \item \textbf{CoLLM-NAS}~\cite{li2026collm}. A collaborative LLM-based NAS framework that uses a stateful Navigator to iteratively refine search strategies from evaluation feedback and historical trajectories, and a stateless Generator to translate these strategies into candidate architectures. A Coordinator manages their interaction, validates generated architectures, and maintains the search archive. Unlike ATHENA, CoLLM-NAS does not leverage cross-site architecture priors.
\end{itemize}

As ATHENA incorporates a two-layer cross-hospital prior comprising the \emph{Layer 1} retrieval prior and the \emph{Layer 2} architecture-effect prior, we conduct ablation studies to quantify the contribution of each component:

\begin{itemize}
    \item \textbf{L1-only.} Retains the retrieval-based cross-hospital prior while removing the architecture-effect prior, quantifying the contribution of \emph{Layer 2}.

    \item \textbf{Leave-One-Task-Out (LOTO) Retrieval.} Excludes task-matched source records during retrieval, forcing knowledge transfer from related but non-identical tasks and evaluating the robustness of retrieval-based priors when exact precedents are unavailable.

    \item \textbf{Cold Start.} Removes both the retrieval prior and the architecture-effect prior, evaluating the effectiveness of the agentic NAS framework in the absence of cross-hospital knowledge.
\end{itemize}

\section{Results}\label{sec_results}

\subsection{Study cohorts}

The cross-hospital prior was constructed from four de-identified OneFlorida+ source sites (Sites A--D). Their pretraining pools comprised 125,819 patients, representing 49.3\% of patients at the four source sites (Table~\ref{tab:cohorts}). The target cohorts were excluded from prior construction. The pretraining pools comprised 54,410 patients at Site E (64.4\% of the cohort) and 42,494 patients in MIMIC-IV (70.0\% of the cohort). The source and target cohorts differed substantially in case mix and data density (\ref{tab:cohort-baseline}~Table in S1 Appendix). Across the six cohorts, Mortality ranged from 0.4\% to 4.2\%, Stay $>$ 7d from 9.6\% to 34.0\%, and Readmission (3M) from 12.7\% to 23.9\%. Median length of stay ranged from 2 to 5 days, and admissions per patient ranged from 1.4 to 2.7. Diagnoses per admission ranged from 5.4 to 19.1 and medications per admission from 3.2 to 13.1. Comorbidity prevalence was also heterogeneous: chronic kidney disease ranged from 2.8\% to 16.0\% of admissions and coronary atherosclerosis from 1.1\% to 23.3\%. Together, these differences provided heterogeneous settings in which to evaluate transfer of architecture knowledge.

\begin{table}[H]
\singlespacing
\footnotesize
\centering
\setlength{\tabcolsep}{2pt}
\renewcommand{\arraystretch}{1.14}
\caption{Pretraining-pool sizes and patient-level fine-tuning splits.}
\label{tab:cohorts}

\centering
\footnotesize
\setlength{\tabcolsep}{2pt}
\renewcommand{\arraystretch}{1.10}

\begin{tabular*}{\columnwidth}{@{\extracolsep{\fill}}lrrrrr@{}}
\toprule
& \multicolumn{2}{c}{Pretraining}
& \multicolumn{3}{c}{Fine-tuning (Binary)} \\
\cmidrule(lr){2-3}\cmidrule(lr){4-6}
Cohort & Patients & Share & Train & Val. & Test \\
\midrule
\multicolumn{6}{@{}l}{\emph{Prior source pool}} \\
Site A & 22,186 & 42.0\% & 6,158 & 12,234 & 12,253 \\
Site B & 29,222 & 52.4\% & 5,300 & 10,566 & 10,647 \\
Site C & 37,152 & 42.0\% & 10,134 & 20,594 & 20,647 \\
Site D & 37,259 & 63.9\% & 4,204 & 8,409 & 8,410 \\
\cmidrule(lr){1-6}
Total & 125,819 & 49.3\% & 25,796 & 51,803 & 51,957 \\
\midrule
\multicolumn{6}{@{}l}{\emph{Target cohorts}} \\
Site E (internal) & 54,410 & 64.4\% & 6,013 & 12,027 & 12,029 \\
MIMIC-IV (external) & 42,494 & 70.0\% & 3,642 & 7,284 & 7,286 \\
\bottomrule
\end{tabular*}
\end{table}

\begin{table}[H]
\ContinuedFloat
\singlespacing
\footnotesize
\centering
\setlength{\tabcolsep}{2pt}
\renewcommand{\arraystretch}{1.10}
\caption{Pretraining-pool sizes and patient-level fine-tuning splits (continued).}

\begin{tabular*}{\columnwidth}{@{\extracolsep{\fill}}llrrr@{}}
\toprule
Cohort & Task & Train & Val. & Test \\
\midrule
\multicolumn{5}{@{}l}{\emph{Prior source pool}} \\
\multirow{3}{*}{Site A}
& Phenotype (6M) & 11,083 & 8,147 & 8,405 \\
& Phenotype (12M) & 12,235 & 9,124 & 9,100 \\
& Drug Recommendation$^{\dagger}$ & 78,999 & 59,249 & 59,251 \\
\addlinespace[1pt]
\multirow{3}{*}{Site B}
& Phenotype (6M) & 9,229 & 6,892 & 6,921 \\
& Phenotype (12M) & 10,285 & 7,792 & 7,643 \\
& Drug Recommendation$^{\dagger}$ & 73,660 & 55,245 & 55,246 \\
\addlinespace[1pt]
\multirow{3}{*}{Site C}
& Phenotype (6M) & 18,257 & 13,922 & 13,699 \\
& Phenotype (12M) & 20,082 & 14,953 & 14,919 \\
& Drug Recommendation$^{\dagger}$ & 92,600 & 69,450 & 69,450 \\
\addlinespace[1pt]
\multirow{3}{*}{Site D}
& Phenotype (6M) & 7,103 & 5,327 & 5,328 \\
& Phenotype (12M) & 7,988 & 5,991 & 5,991 \\
& Drug Recommendation$^{\dagger}$ & 22,958 & 17,218 & 17,219 \\
\cmidrule(lr){1-5}
\multirow{3}{*}{Total}
& Phenotype (6M) & 45,672 & 34,288 & 34,353 \\
& Phenotype (12M) & 50,590 & 37,860 & 37,653 \\
& Drug Recommendation$^{\dagger}$ & 268,217 & 201,162 & 201,166 \\
\midrule
\multicolumn{5}{@{}l}{\emph{Target cohorts}} \\
\multirow{3}{*}{Site E}
& Phenotype (6M) & 9,195 & 6,896 & 6,897 \\
& Phenotype (12M) & 10,382 & 7,786 & 7,787 \\
& Drug Recommendation$^{\dagger}$ & 33,380 & 25,035 & 25,036 \\
\addlinespace[1pt]
\multirow{3}{*}{MIMIC-IV}
& Phenotype (6M) & 3,866 & 2,900 & 2,901 \\
& Phenotype (12M) & 4,478 & 3,359 & 3,360 \\
& Drug Recommendation$^{\dagger}$ & 24,017 & 18,012 & 18,014 \\
\bottomrule
\end{tabular*}

\par\smallskip
{\footnotesize OneFlorida+ contributor sites are de-identified as Site~A--E. \emph{Share} denotes the fraction of cohort patients in the pretraining pool. Fine-tuning counts are shown in separate training, validation, and test columns. \emph{Binary} denotes the common split for Mortality, Stay $>$ 7d, and Readmission (3M). Phenotype and Drug Recommendation are abbreviated as Pheno. and Drug Rec., respectively. $^{\dagger}$Drug Recommendation uses a separately defined drug-eligible cohort ($\geq 1$ in-vocabulary medication) and an independent patient-level split.\par}
\end{table}

\subsection{Supernet ranking fidelity and search efficiency}

Before using supernet-based evaluations to guide architecture search, we first assessed whether weight-inherited subnetworks preserved the relative performance ranking of independently trained architectures. For each MIMIC-IV task, we sampled 150 valid Transformer architectures and evaluated each under two settings: fine-tuning with weights inherited from the pretrained supernet and independent pretraining followed by fine-tuning. We then computed Spearman rank correlations between the resulting architecture rankings based on AUROC and AUPRC. This analysis examined whether the computationally cheaper supernet-based approach could serve as a reliable proxy for conventional pretrain-then-fine-tune evaluation.

The AutoFormer-style weight-sharing supernet preserved the relative performance of conventionally pretrained and fine-tuned Transformer architectures (Table~\ref{tab:regression}). Spearman correlations were positive for every task and metric. Ranking fidelity was particularly high for the three multilabel tasks: correlations ranged from 0.940 to 0.951 for Phenotype (6M) and Phenotype (12M), and from 0.944 to 0.958 for Drug Recommendation. Stay $>$ 7d also showed strong agreement ($\rho=0.809$ for AUROC and $\rho=0.806$ for AUPRC). Agreement was more moderate for Mortality and Readmission (3M), with the lowest correlation observed for Readmission (3M) AUPRC ($\rho=0.412$). Thus, although fidelity varied by task, the supernet provided an informative proxy ranking among the sampled architectures.

\begin{table}[H]
\singlespacing
\footnotesize
\centering
\setlength{\tabcolsep}{2pt}
\renewcommand{\arraystretch}{1.14}
\caption{Supernet ranking fidelity on MIMIC-IV.}
\label{tab:regression}
\begin{tabular*}{\tblwidth}{@{}LCC@{}}
\toprule
Task & $\rho$ (AUROC) & $\rho$ (AUPRC) \\
\midrule
Mortality           & 0.544 & 0.567 \\
Stay $>$ 7d         & 0.809 & 0.806 \\
Readmission (3M)    & 0.570 & 0.412 \\
Phenotype (6M)      & 0.951 & 0.944 \\
Phenotype (12M)     & 0.940 & 0.944 \\
Drug Recommendation & 0.944 & 0.958 \\
\bottomrule
\end{tabular*}

\par\smallskip
{\footnotesize For each task, 150 architectures are sampled and evaluated both as weight-inherited subnetworks of the pretrained supernet and through independent pretraining followed by fine-tuning. $\rho$ denotes the Spearman rank correlation between the resulting architecture rankings.\par}
\end{table}

We next measured the graphics processing unit (GPU) time required to evaluate increasing numbers of candidate architectures on one NVIDIA L4. The weight-sharing calculation included one supernet pretraining run followed by subnet fine-tuning for each candidate, whereas conventional search repeated pretraining and fine-tuning independently for every candidate. Weight sharing yielded larger reductions in search cost as more architectures were evaluated (Table~\ref{tab:compute}). At a budget of five evaluations, the measured speedup over conventional pretrain-then-fine-tune evaluation was 4.0$\times$ on OneFlorida+ and 4.4$\times$ on MIMIC-IV. At 30 evaluations, the corresponding speedups increased to 10.7$\times$ and 15.3$\times$, reducing the estimated cost from 960 to 90 GPU-minutes on OneFlorida+ and from 900 to 59 GPU-minutes on MIMIC-IV. 

\begin{table}[H]
\singlespacing
\footnotesize
\centering
\setlength{\tabcolsep}{2pt}
\renewcommand{\arraystretch}{1.14}
\caption{Search compute under weight sharing, in GPU-minutes on one NVIDIA L4.}
\label{tab:compute}
\begin{tabular*}{\tblwidth}{@{}CCCCCCC@{}}
\toprule
& \multicolumn{3}{c}{OneFlorida+ held-out} & \multicolumn{3}{c}{MIMIC-IV} \\
\cmidrule(lr){2-4}\cmidrule(lr){5-7}
$N$ & Shared & Scratch & Speedup & Shared & Scratch & Speedup \\
\midrule
5  & 40 & 160 & 4.0$\times$  & 34 & 150 & 4.4$\times$ \\
10 & 50 & 320 & 6.4$\times$  & 39 & 300 & 7.7$\times$ \\
20 & 70 & 640 & 9.1$\times$  & 49 & 600 & 12.2$\times$ \\
30 & \textbf{90} & \textbf{960} & \textbf{10.7$\times$} & \textbf{59} & \textbf{900} & \textbf{15.3$\times$} \\
\bottomrule
\end{tabular*}

\par\smallskip
{\footnotesize The shared-weight approach pretrains the supernet \emph{once} and fine-tunes each of $N$ candidates on it ($T_{\mathrm{pre}}+N\,T_{\mathrm{ft}}$); conventional from-scratch NAS incurs $N(T_{\mathrm{pre}}+T_{\mathrm{ft}})$. Values at the maximum budget of 30 evaluations are shown in \textbf{bold}.\par}
\end{table}

\subsection{Task-specific architecture-effect prior}

The cross-hospital architecture-effect prior captured both shared and task-specific Transformer design preferences across the six clinical tasks (\ref{fig:supp-prior}~Fig in S1 Appendix). The most consistent signal was model width. Embedding dimensions of 128 and 256 were reliably preferred for Mortality, while dimensions of 64, 128, and 256 were preferred for the other five tasks. Increasing the number of attention heads was also generally favorable: configurations with 2, 4, or 8 heads had positive effects for Mortality, Stay $>$ 7d, Readmission (3M), Phenotype (6M), and Phenotype (12M). For Drug Recommendation, 4 and 8 heads had positive effects relative to the one-head reference level, whereas the effect of 2 heads was inconclusive.

Depth showed a more heterogeneous pattern. Mortality and Stay $>$ 7d favored depths of 2, 4, and 8 over the single-layer reference configuration. Readmission (3M) did not show the same trend: depths of 2 and 8 were discouraged, while the effect of a depth of 4 was inconclusive. Among the multilabel tasks, Phenotype (6M) favored a depth of 2 but discouraged depths of 4 and 8; Phenotype (12M) and Drug Recommendation discouraged all depths greater than one. The MLP expansion ratio showed a similarly conservative pattern. Ratios greater than one were consistently discouraged for Mortality, Stay $>$ 7d, Readmission (3M), and Phenotype (6M). For Phenotype (12M), ratios of 4 and 8 were discouraged and a ratio of 2 was inconclusive; for Drug Recommendation, a ratio of 4 was discouraged while ratios of 2 and 8 were inconclusive. Thus, the prior identified wider embeddings and multi-head attention as relatively transferable signals, while retaining task-dependent guidance for depth and MLP expansion ratio rather than prescribing one architecture for all outcomes.

Pairwise rules derived from each task's two most influential features were also supplied to the search prompt (\ref{tab:interaction-rules}~Table in S1 Appendix). For five of the six tasks, the direction of the identified preference remained unchanged across all levels of the second feature. Mortality was the exception, where the preference for an embedding dimension of 256 emerged when the number of attention heads was at least two.

\subsection{Predictive performance under different search budgets}

We evaluated ATHENA's predictive performance in two complementary settings. First, we conducted a strict equal-compute comparison in which ATHENA was run with a search budget of 30 architecture evaluations and each baseline followed the conventional paradigm of independently pretraining and then fine-tuning each candidate under a total compute budget matched to ATHENA. Second, to isolate the quality of the search policy, all methods started from the same pretrained AutoFormer supernet and differed only in how they searched the architecture space.

Under the strict equal-compute setting (Table~\ref{tab:scratch-proxy}), ATHENA achieved the highest mean test AUPRC in 10 of the 12 hospital--task evaluations and significantly outperformed every baseline in 9 of 12. On OneFlorida+, ATHENA ranked first in four of six tasks and significantly outperformed all baselines for Stay $>$ 7d, Phenotype (12M), and Drug Recommendation. On MIMIC-IV, it ranked first and significantly outperformed every baseline across all six tasks. Notably, the conventional baselines evaluated each candidate through independent pretraining and fine-tuning, providing direct performance estimates without the approximation error associated with weight-sharing proxy evaluations, but at a substantially higher cost per candidate. Despite this trade-off, ATHENA retained its performance advantage when all methods were constrained to the same total compute budget.

\begin{table}[H]
\singlespacing
\footnotesize
\centering
\setlength{\tabcolsep}{2pt}
\renewcommand{\arraystretch}{1.14}
\caption{Test AUPRC across target hospitals and prediction tasks under an equal-compute comparison between ATHENA and baselines following the conventional pretrain-then-finetune paradigm, with all methods evaluated under a compute budget matched to that of ATHENA's search with a maximum budget of 30 architecture evaluations (mean $\pm$ standard deviation (SD) over five random seeds).}
\label{tab:scratch-proxy}
\footnotesize
\begin{tabular*}{\linewidth}{@{\extracolsep{\fill}}p{0.68in}p{1.02in}*{5}{>{\centering\arraybackslash}p{0.86in}}@{}}
\toprule
Target & Task & Random & EA & GENIUS & CoLLM-NAS & ATHENA \\
\midrule
OneFlorida+ & Mortality
& \underline{58.95 $\pm$ 1.01}
& \underline{58.95 $\pm$ 1.01}
& \textbf{59.38 $\pm$ 0.43}
& 58.86 $\pm$ 2.07
& 58.85 $\pm$ 0.84 \\

OneFlorida+ & Stay $>$ 7d
& \underline{71.76 $\pm$ 2.31}
& \underline{71.76 $\pm$ 2.31}
& 71.56 $\pm$ 0.33
& 71.16 $\pm$ 4.24
& \textbf{73.90 $\pm$ 0.27}$^{*}$ \\

OneFlorida+ & Readmission (3M)
& 53.18 $\pm$ 0.38
& 53.18 $\pm$ 0.38
& \textbf{53.45 $\pm$ 0.43}
& \underline{53.20 $\pm$ 0.28}
& 52.56 $\pm$ 0.24 \\

OneFlorida+ & Phenotype (6M)
& 29.42 $\pm$ 2.72
& 29.42 $\pm$ 2.72
& 29.76 $\pm$ 0.29
& \underline{30.48 $\pm$ 0.96}
& \textbf{31.06 $\pm$ 0.55} \\

OneFlorida+ & Phenotype (12M)
& 30.96 $\pm$ 2.37
& 30.96 $\pm$ 2.37
& 31.46 $\pm$ 0.14
& \underline{31.70 $\pm$ 0.81}
& \textbf{32.54 $\pm$ 0.40}$^{*}$ \\

OneFlorida+ & Drug Recommendation
& 18.91 $\pm$ 0.54
& 18.91 $\pm$ 0.54
& 19.06 $\pm$ 0.15
& \underline{19.08 $\pm$ 0.25}
& \textbf{19.34 $\pm$ 0.07}$^{*}$ \\

\cmidrule(lr){1-7}

MIMIC-IV & Mortality
& 78.68 $\pm$ 0.89
& 78.68 $\pm$ 0.89
& \underline{78.78 $\pm$ 0.93}
& 78.67 $\pm$ 1.07
& \textbf{79.75 $\pm$ 0.37}$^{*}$ \\

MIMIC-IV & Stay $>$ 7d
& 73.32 $\pm$ 1.77
& 73.32 $\pm$ 1.77
& \underline{75.31 $\pm$ 0.65}
& 75.07 $\pm$ 0.78
& \textbf{75.88 $\pm$ 0.43}$^{*}$ \\

MIMIC-IV & Readmission (3M)
& 44.28 $\pm$ 0.46
& 44.28 $\pm$ 0.46
& \underline{44.39 $\pm$ 0.32}
& 43.99 $\pm$ 0.43
& \textbf{45.13 $\pm$ 0.53}$^{*}$ \\

MIMIC-IV & Phenotype (6M)
& 48.03 $\pm$ 2.88
& 48.03 $\pm$ 2.88
& \underline{48.42 $\pm$ 0.24}
& 48.02 $\pm$ 0.33
& \textbf{49.84 $\pm$ 0.35}$^{*}$ \\

MIMIC-IV & Phenotype (12M)
& 47.95 $\pm$ 1.88
& 47.95 $\pm$ 1.88
& \underline{48.07 $\pm$ 0.61}
& 47.94 $\pm$ 0.33
& \textbf{49.42 $\pm$ 0.20}$^{*}$ \\

MIMIC-IV & Drug Recommendation
& 20.25 $\pm$ 1.05
& 20.25 $\pm$ 1.05
& \underline{20.34 $\pm$ 0.20}
& 20.07 $\pm$ 0.21
& \textbf{21.08 $\pm$ 0.08}$^{*}$ \\
\bottomrule
\end{tabular*}

\par\smallskip
{\footnotesize Paired seed-level bootstrap (1,000 resamples; five matched seeds; fixed test set) yielded one-sided 97.5\% lower confidence bounds for ATHENA--baseline AUPRC differences. $^*$ indicates a positive bound against all baselines (intersection--union test, $\alpha=0.025$; no baseline-wise multiplicity adjustment). Bold and underlined values denote the highest and second-highest row means (ties included), regardless of significance.\par}
\end{table}

When all methods searched the same pretrained AutoFormer supernet, ATHENA achieved the best overall test AUPRC ranking across all search budgets (Table~\ref{tab:main-auprc}). With only five architecture evaluations, it ranked first in 10 of 12 hospital--task comparisons and second in the remaining two, yielding an average rank of 1.17. It ranked first across all six OneFlorida+ tasks and four of six MIMIC-IV tasks, including all multilabel tasks at both targets. The largest gains over the strongest baseline were observed for OneFlorida+ Phenotype (12M), OneFlorida+ Phenotype (6M), and MIMIC-IV Drug Recommendation.

\clearpage
\begingroup
\singlespacing
\footnotesize
\setlength{\tabcolsep}{2pt}
\renewcommand{\arraystretch}{1.14}
\begin{longtable}{@{}p{0.68in}p{1.02in}*{5}{>{\centering\arraybackslash}p{0.86in}}@{}}
\caption{Test AUPRC across target hospitals and prediction tasks under increasing NAS budgets of 5, 20, and 30 architecture evaluations (mean $\pm$ standard deviation over five random seeds).}\label{tab:main-auprc}\\
\toprule
Target & Task & Random Search & EA & GENIUS & CoLLM-NAS & ATHENA \\
\midrule
\endfirsthead
\caption[]{Test AUPRC across target hospitals and prediction tasks under increasing NAS budgets of 5, 20, and 30 architecture evaluations (mean $\pm$ standard deviation over five random seeds). (continued)}\\
\toprule
Target & Task & Random Search & EA & GENIUS & CoLLM-NAS & ATHENA \\
\midrule
\endhead
\midrule
\multicolumn{7}{r}{\footnotesize Continued on next page}\\
\endfoot
\endlastfoot

\multicolumn{7}{c}{\textit{(a) Search budget = 5 evaluations}} \\
\midrule
OneFlorida+ & Mortality & 57.78 $\pm$ 0.90 & 57.78 $\pm$ 0.90 & \underline{58.04 $\pm$ 0.34} & 57.72 $\pm$ 1.36 & \textbf{58.23 $\pm$ 1.72} \\
OneFlorida+ & Stay $>$ 7d & 73.25 $\pm$ 0.33 & 73.25 $\pm$ 0.33 & \underline{74.00 $\pm$ 0.34} & 73.59 $\pm$ 0.45 & \textbf{74.03 $\pm$ 0.37} \\
OneFlorida+ & Readmission (3M) & 52.09 $\pm$ 0.42 & 52.09 $\pm$ 0.42 & 52.06 $\pm$ 0.58 & \underline{52.16 $\pm$ 0.40} & \textbf{52.26 $\pm$ 0.27} \\
OneFlorida+ & Phenotype (6M) & \underline{29.65 $\pm$ 0.59} & 29.53 $\pm$ 0.40 & 29.28 $\pm$ 0.87 & 29.29 $\pm$ 1.50 & \textbf{30.42 $\pm$ 0.51} \\
OneFlorida+ & Phenotype (12M) & \underline{30.86 $\pm$ 0.71} & 30.74 $\pm$ 0.69 & 30.39 $\pm$ 0.57 & 30.23 $\pm$ 0.60 & \textbf{32.17 $\pm$ 0.42} \\
OneFlorida+ & Drug Recommendation & 19.09 $\pm$ 0.24 & 19.09 $\pm$ 0.24 & 19.11 $\pm$ 0.27 & \underline{19.16 $\pm$ 0.22} & \textbf{19.32 $\pm$ 0.08} \\
\cmidrule(lr){1-7}
MIMIC-IV & Mortality & 79.25 $\pm$ 0.64 & 79.25 $\pm$ 0.64 & 79.29 $\pm$ 0.62 & \textbf{79.52 $\pm$ 0.43} & \underline{79.31 $\pm$ 0.40} \\
MIMIC-IV & Stay $>$ 7d & 74.67 $\pm$ 0.41 & 74.80 $\pm$ 0.59 & \underline{75.22 $\pm$ 0.66} & 75.02 $\pm$ 0.60 & \textbf{75.66 $\pm$ 0.71} \\
MIMIC-IV & Readmission (3M) & 43.64 $\pm$ 0.47 & 43.64 $\pm$ 0.47 & 44.29 $\pm$ 0.26 & \textbf{44.55 $\pm$ 0.55} & \underline{44.47 $\pm$ 0.44} \\
MIMIC-IV & Phenotype (6M) & 48.96 $\pm$ 0.76 & \underline{48.97 $\pm$ 0.76} & 48.19 $\pm$ 0.66 & 48.63 $\pm$ 0.67 & \textbf{49.25 $\pm$ 0.57} \\
MIMIC-IV & Phenotype (12M) & 48.64 $\pm$ 0.41 & 48.63 $\pm$ 0.39 & 47.98 $\pm$ 0.41 & \underline{48.79 $\pm$ 0.64} & \textbf{49.26 $\pm$ 0.09} \\
MIMIC-IV & Drug Recommendation & \underline{20.63 $\pm$ 0.43} & \underline{20.63 $\pm$ 0.43} & 20.36 $\pm$ 0.36 & 20.24 $\pm$ 0.12 & \textbf{21.16 $\pm$ 0.17} \\
\cmidrule(lr){1-7}
\multicolumn{2}{@{}l}{\textit{Avg.\ Rank (best-in-row)}} & 3.54 (0) & 3.62 (0) & 3.58 (0) & \underline{3.08 (2)} & \textbf{1.17 (10)} \\
\midrule
\multicolumn{7}{c}{\textit{(b) Search budget = 20 evaluations}} \\
\midrule
OneFlorida+ & Mortality & 57.46 $\pm$ 0.50 & 58.10 $\pm$ 0.48 & \underline{58.18 $\pm$ 0.61} & 57.88 $\pm$ 0.92 & \textbf{58.85 $\pm$ 0.84} \\
OneFlorida+ & Stay $>$ 7d & 73.28 $\pm$ 0.52 & 73.42 $\pm$ 0.23 & 73.60 $\pm$ 0.73 & \underline{73.65 $\pm$ 0.51} & \textbf{73.90 $\pm$ 0.28} \\
OneFlorida+ & Readmission (3M) & 52.21 $\pm$ 0.22 & 52.34 $\pm$ 0.18 & 52.40 $\pm$ 0.16 & \textbf{52.49 $\pm$ 0.12} & \underline{52.44 $\pm$ 0.28} \\
OneFlorida+ & Phenotype (6M) & \underline{30.57 $\pm$ 0.20} & 30.23 $\pm$ 1.00 & 30.02 $\pm$ 0.52 & 29.46 $\pm$ 1.57 & \textbf{31.06 $\pm$ 0.55} \\
OneFlorida+ & Phenotype (12M) & 31.53 $\pm$ 0.66 & \underline{32.14 $\pm$ 0.85} & 31.37 $\pm$ 0.76 & 30.54 $\pm$ 1.06 & \textbf{32.62 $\pm$ 0.29} \\
OneFlorida+ & Drug Recommendation & 19.28 $\pm$ 0.12 & 19.27 $\pm$ 0.21 & \textbf{19.44 $\pm$ 0.06} & 19.29 $\pm$ 0.19 & \underline{19.32 $\pm$ 0.04} \\
\cmidrule(lr){1-7}
MIMIC-IV & Mortality & 79.16 $\pm$ 0.41 & 79.46 $\pm$ 0.24 & 79.21 $\pm$ 0.45 & \underline{79.48 $\pm$ 0.63} & \textbf{79.75 $\pm$ 0.37} \\
MIMIC-IV & Stay $>$ 7d & 75.18 $\pm$ 0.64 & \textbf{75.67 $\pm$ 0.26} & 75.39 $\pm$ 0.45 & 75.20 $\pm$ 0.58 & \underline{75.64 $\pm$ 0.48} \\
MIMIC-IV & Readmission (3M) & 44.25 $\pm$ 0.35 & \underline{44.88 $\pm$ 0.97} & 44.61 $\pm$ 0.46 & \textbf{44.94 $\pm$ 0.57} & \textbf{44.94 $\pm$ 0.55} \\
MIMIC-IV & Phenotype (6M) & \underline{49.43 $\pm$ 0.53} & 49.22 $\pm$ 0.69 & 49.26 $\pm$ 0.69 & 49.01 $\pm$ 0.73 & \textbf{49.84 $\pm$ 0.35} \\
MIMIC-IV & Phenotype (12M) & \underline{49.10 $\pm$ 0.16} & 49.04 $\pm$ 0.28 & 49.05 $\pm$ 0.24 & 48.66 $\pm$ 0.59 & \textbf{49.42 $\pm$ 0.20} \\
MIMIC-IV & Drug Recommendation & \underline{21.08 $\pm$ 0.09} & 20.98 $\pm$ 0.33 & 20.75 $\pm$ 0.35 & 20.39 $\pm$ 0.32 & \textbf{21.15 $\pm$ 0.13} \\
\cmidrule(lr){1-7}
\multicolumn{2}{@{}l}{\textit{Avg.\ Rank (best-in-row)}} & 3.75 (0) & 3.25 (1) & \underline{3.17 (1)} & 3.54 (2) & \textbf{1.29 (9)} \\
\midrule
\multicolumn{7}{c}{\textit{(c) Search budget = 30 evaluations}} \\
\midrule
OneFlorida+ & Mortality & 57.91 $\pm$ 0.40 & 58.07 $\pm$ 0.49 & \underline{58.18 $\pm$ 0.61} & 57.88 $\pm$ 0.92 & \textbf{58.85 $\pm$ 0.84} \\
OneFlorida+ & Stay $>$ 7d & 73.52 $\pm$ 0.78 & \underline{73.97 $\pm$ 0.40} & \textbf{74.10 $\pm$ 0.13} & 73.65 $\pm$ 0.51 & 73.90 $\pm$ 0.27 \\
OneFlorida+ & Readmission (3M) & 52.27 $\pm$ 0.24 & 52.41 $\pm$ 0.20 & \underline{52.54 $\pm$ 0.23} & 52.49 $\pm$ 0.12 & \textbf{52.56 $\pm$ 0.24} \\
OneFlorida+ & Phenotype (6M) & 30.57 $\pm$ 0.20 & \underline{31.01 $\pm$ 0.30} & 30.02 $\pm$ 0.52 & 29.46 $\pm$ 1.57 & \textbf{31.06 $\pm$ 0.55} \\
OneFlorida+ & Phenotype (12M) & 31.71 $\pm$ 0.49 & \underline{32.34 $\pm$ 0.75} & 31.44 $\pm$ 0.71 & 30.54 $\pm$ 1.06 & \textbf{32.54 $\pm$ 0.40} \\
OneFlorida+ & Drug Recommendation & 19.32 $\pm$ 0.11 & \underline{19.43 $\pm$ 0.08} & \textbf{19.44 $\pm$ 0.06} & 19.29 $\pm$ 0.19 & 19.34 $\pm$ 0.07 \\
\cmidrule(lr){1-7}
MIMIC-IV & Mortality & 79.16 $\pm$ 0.42 & 79.30 $\pm$ 0.54 & 79.12 $\pm$ 0.57 & \underline{79.48 $\pm$ 0.63} & \textbf{79.75 $\pm$ 0.37} \\
MIMIC-IV & Stay $>$ 7d & 75.01 $\pm$ 0.50 & \underline{75.69 $\pm$ 0.41} & 75.56 $\pm$ 0.42 & 75.20 $\pm$ 0.58 & \textbf{75.88 $\pm$ 0.43} \\
MIMIC-IV & Readmission (3M) & 44.76 $\pm$ 0.16 & 44.78 $\pm$ 1.02 & 44.79 $\pm$ 0.33 & \underline{44.94 $\pm$ 0.57} & \textbf{45.13 $\pm$ 0.53} \\
MIMIC-IV & Phenotype (6M) & \underline{49.61 $\pm$ 0.51} & 49.56 $\pm$ 0.48 & 49.22 $\pm$ 0.68 & 49.01 $\pm$ 0.73 & \textbf{49.84 $\pm$ 0.35} \\
MIMIC-IV & Phenotype (12M) & 49.11 $\pm$ 0.14 & \underline{49.14 $\pm$ 0.29} & 49.05 $\pm$ 0.24 & 48.66 $\pm$ 0.59 & \textbf{49.42 $\pm$ 0.20} \\
MIMIC-IV & Drug Recommendation & 21.03 $\pm$ 0.12 & \textbf{21.15 $\pm$ 0.06} & 20.78 $\pm$ 0.39 & 20.39 $\pm$ 0.32 & \underline{21.08 $\pm$ 0.08} \\
\cmidrule(lr){1-7}
\multicolumn{2}{@{}l}{\textit{Avg.\ Rank (best-in-row)}} & 3.83 (0) & \underline{2.50 (1)} & 3.08 (2) & 4.17 (0) & \textbf{1.42 (9)} \\
\bottomrule
\end{longtable}
\noindent\begin{minipage}{\linewidth}

{\footnotesize For each row, the best mean is in \textbf{bold} and the second-best mean is \underline{underlined}; tied values share the mark. \textit{Avg.\ Rank} is the mean within-row rank over the 12 hospital--task comparisons in each panel (1 = best; ties receive the average rank), with the number of best-in-row comparisons in parentheses. All methods search the same pretrained supernet and therefore differ only in search strategy.\par}

\end{minipage}
\endgroup

This advantage persisted at larger budgets. At both 20 and 30 evaluations, ATHENA ranked first in 9 of 12 comparisons, with average ranks of 1.29 and 1.42, respectively, compared with 3.17 and 2.50 for the next-best baseline. At budget 30, the AUROC analysis (\ref{tab:auroc}~Table in S1 Appendix) showed a similar pattern, with ATHENA ranking first in 7 of 12 comparisons (including one tie with EA) and achieving the best average rank (1.71). Overall, the advantage was consistent across both target health systems.

\begin{figure}[H]
\centering
\includegraphics[width=\linewidth,height=0.72\textheight,keepaspectratio]{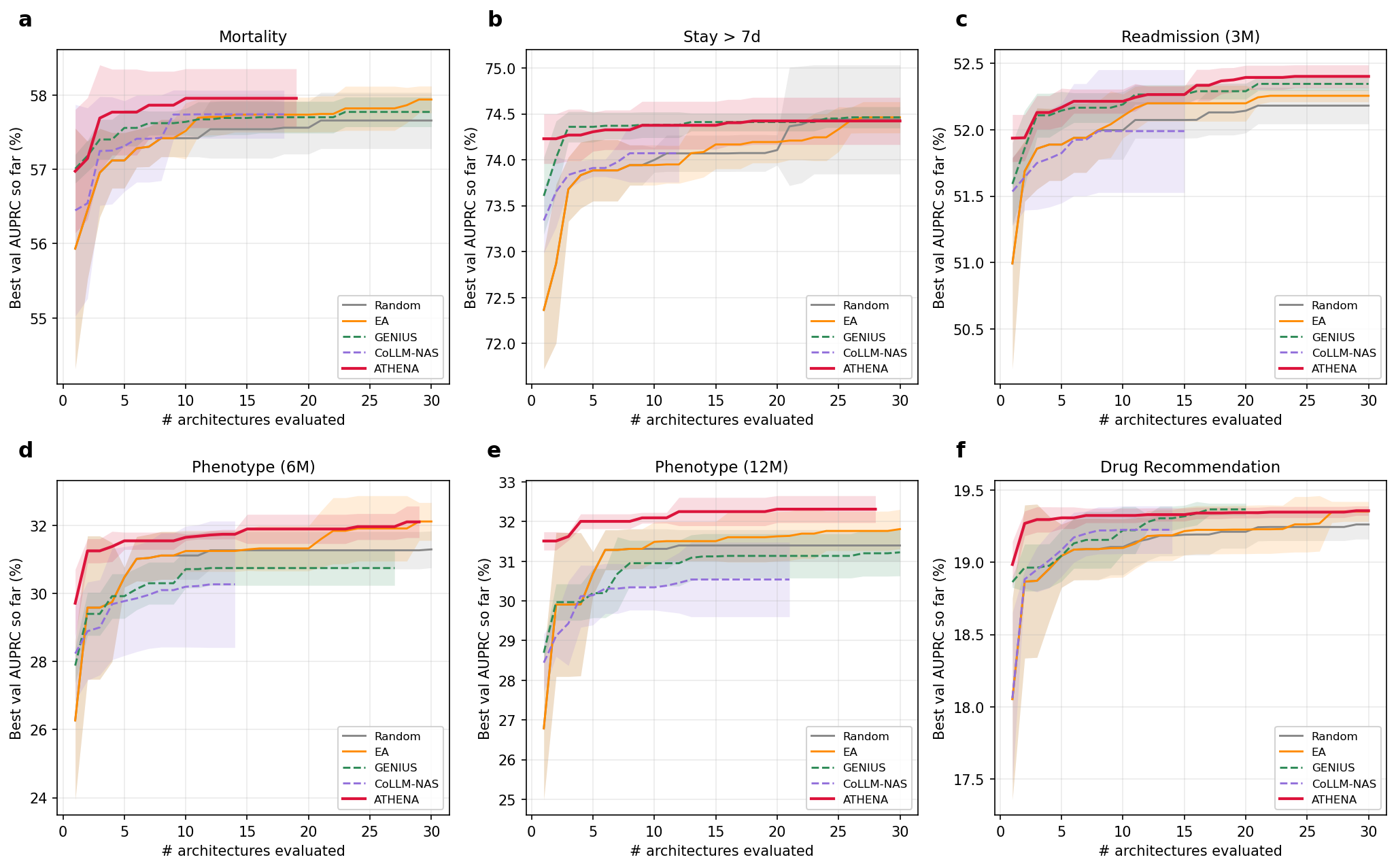}
\caption{Search trajectories on the held-out OneFlorida+ target. Panels show (\textbf{a}) Mortality, (\textbf{b}) Stay $>$ 7d, (\textbf{c}) Readmission (3M), (\textbf{d}) Phenotype (6M), (\textbf{e}) Phenotype (12M), and (\textbf{f}) Drug Recommendation. Lines and shaded bands show the mean and SD, respectively, of the best validation AUPRC attained across five random seeds as a function of the number of evaluated architectures. Curves may end before the full budget when proposal saturation yields no new legal architecture.}
\label{fig:search-trajectory-source15}
\end{figure}

Validation trajectories further showed that ATHENA's gains generally emerged within the first few evaluations rather than only after most of the search budget had been consumed (Fig~\ref{fig:search-trajectory-source15}; \ref{fig:supp-search-trajectory-mimic}~Fig in S1 Appendix). This pattern was most evident for Phenotype (6M), Phenotype (12M), and Drug Recommendation, for which ATHENA rapidly attained strong validation AUPRC and remained leading or competitive through most of the search. The margins were most persistent for Phenotype (12M) on both targets, whereas competing methods narrowed or occasionally closed the gap on the other tasks as the budget increased. Mortality, Stay $>$ 7d, and Readmission (3M) showed closer convergence among methods. Some LLM-based searches terminated before 30 evaluations when they could no longer generate new legal, non-duplicate candidates. Overall, ATHENA retained most of its early gains, with smaller and task-dependent improvements from later evaluations, indicating that much of its advantage was established under a relatively small search budget.

\subsection{Performance--complexity trade-offs and selection behavior}

\begin{figure}[H]
\centering
\includegraphics[width=\linewidth,height=0.72\textheight,keepaspectratio]{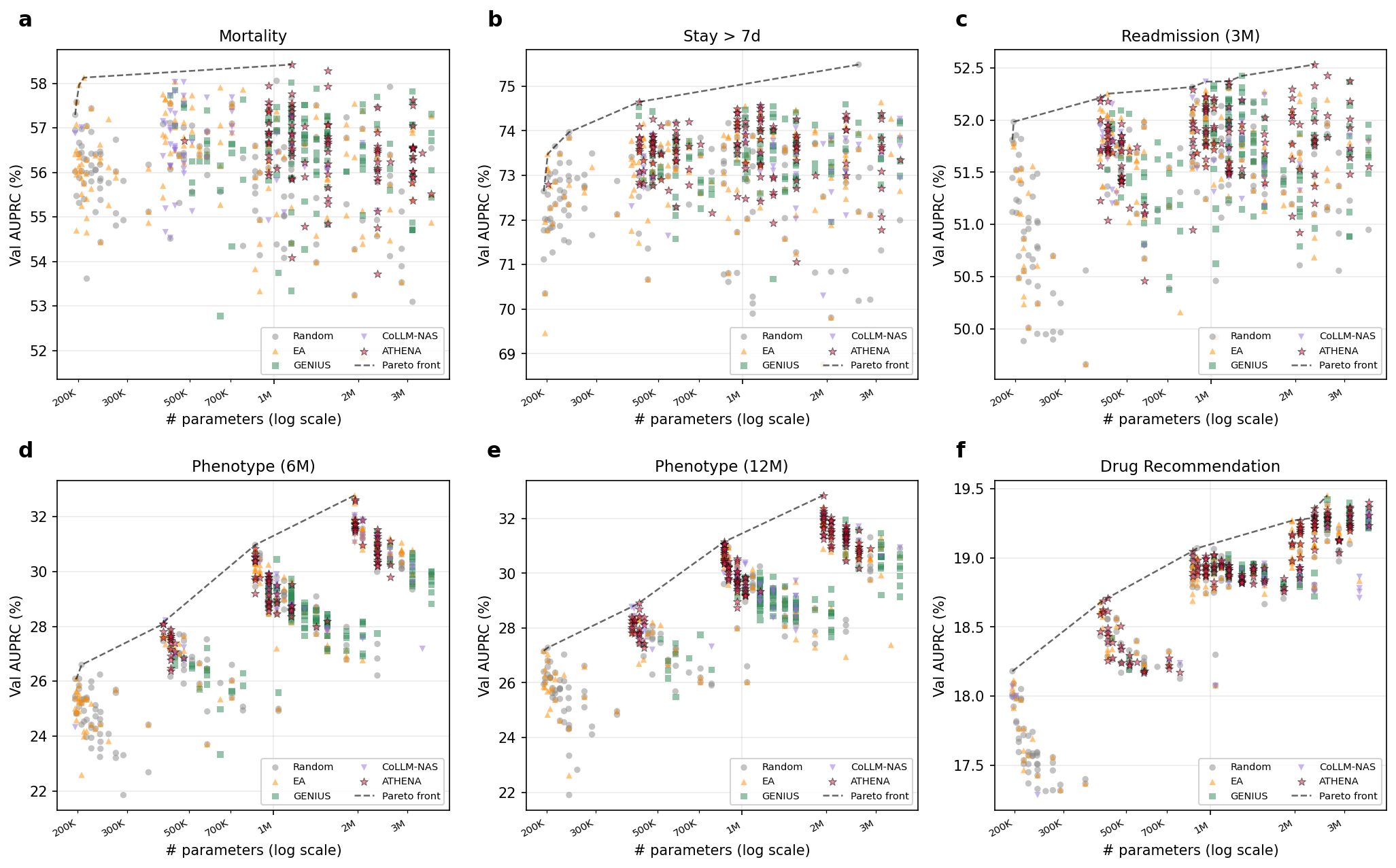}
\caption{Validation AUPRC versus parameter count on the held-out OneFlorida+ target. Markers identify the search method and the dashed line denotes the empirical Pareto frontier across evaluated architectures.}
\label{fig:pareto-source15}
\end{figure}

The validation Pareto analyses showed that performance was not a monotonic function of model size (Fig~\ref{fig:pareto-source15}; \ref{fig:supp-pareto-mimic}~Fig in S1 Appendix). In the binary tasks, relatively small architectures could be competitive with substantially larger models, while architectures with similar parameter counts often had visibly different AUPRC values. The multilabel tasks exhibited stepwise improvements across parameter bands, but still showed considerable within-band variation. ATHENA evaluated candidates across these bands and repeatedly placed architectures near the empirical Pareto frontier rather than concentrating exclusively at the largest end of the search space. This pattern was present on both the internal and external targets.

ATHENA's predictive performance was not explained by systematically selecting the largest models or by using more candidate evaluations (Table~\ref{tab:behavior}). On OneFlorida+, the architectures selected by ATHENA averaged 1.55 million parameters, compared with 1.90 million for GENIUS, despite ATHENA's substantially better average performance rank. ATHENA evaluated an average of 22.5 architectures on OneFlorida+ and 19.6 on MIMIC-IV, below the maximum budget of 30 and slightly fewer than GENIUS on both targets.

\begin{table}[H]
\singlespacing
\footnotesize
\centering
\setlength{\tabcolsep}{2pt}
\renewcommand{\arraystretch}{1.14}
\caption{Search behavior at a budget of 30 evaluations, averaged over the six tasks and five random seeds.}
\label{tab:behavior}
\begin{tabular*}{\tblwidth}{@{}LCCCC@{}}
\toprule
& \multicolumn{2}{c}{OneFlorida+} & \multicolumn{2}{c}{MIMIC-IV} \\
\cmidrule(lr){2-3}\cmidrule(lr){4-5}
Method & Params (M) & Evals & Params (M) & Evals \\
\midrule
Random Search   & 1.28 & 30.0 & 2.34 & 30.0 \\
EA              & 1.33 & 30.0 & 2.11 & 30.0 \\
GENIUS          & 1.90 & 23.6 & 2.42 & 20.3 \\
CoLLM-NAS       & 1.28 & 10.7 & 1.93 & 9.1 \\
ATHENA          & 1.55 & 22.5 & 2.58 & 19.6 \\
\bottomrule
\end{tabular*}

\par\smallskip
{\footnotesize \emph{Params} is the size of the selected architecture in millions of parameters; \emph{Evals} is the number of distinct architectures actually evaluated.\par}
\end{table}

Architecture selection was also more reproducible across repeated searches (Table~\ref{tab:arch-config}). For each task $t$, we calculated the modal architecture selection rate as the proportion of the five random seeds selecting the most frequently chosen architecture, and averaged this rate across the six tasks:
\begin{equation}
\label{eq:modal-selection-rate}
\text{Mean modal selection (\%)} =
\frac{1}{6}\sum_{t=1}^{6}
100\times\frac{\max_a(n_{t,a})}{5},
\end{equation}
where $n_{t,a}$ is the number of random seeds selecting architecture $a$ for task $t$. On OneFlorida+, ATHENA and EA achieved the highest mean modal architecture selection rate of 40.0\%, compared with 26.7\% for Random Search and 33.3\% for both GENIUS and CoLLM-NAS. On MIMIC-IV, ATHENA achieved the highest rate at 43.3\%, compared with 20.0\%--33.3\% for the baselines. Higher rates indicate greater convergence toward the same architecture across repeated searches and reflect selection consistency rather than architecture quality.

\begin{table}[H]
\singlespacing
\footnotesize
\centering
\setlength{\tabcolsep}{2pt}
\renewcommand{\arraystretch}{1.14}
\caption{Modal architecture selected by each method, reported as the tuple (embed\_dim, depth, heads, mlp\_ratio), with the number of random seeds selecting that architecture (out of five) in parentheses.}
\label{tab:arch-config}
\footnotesize
\begin{tabular*}{\linewidth}{@{\extracolsep{\fill}}p{0.68in}p{1.02in}*{5}{>{\centering\arraybackslash}p{0.86in}}@{}}
\toprule
Target & Task & Random Search & EA & GENIUS & CoLLM-NAS & ATHENA \\
\midrule
OneFlorida+ & Mortality & \textemdash & (64, 2, 2, 2) (2) & (256, 2, 8, 1) (2) & \textemdash & \textemdash \\
OneFlorida+ & Stay $>$ 7d & \textemdash & \textemdash & \textemdash & \textemdash & (128, 4, 4, 1) (2) \\
OneFlorida+ & Readmission (3M) & (128, 1, 8, 1) (2) & \textemdash & \textemdash & (128, 4, 4, 2) (2) & (128, 2, 8, 2) (2) \\
OneFlorida+ & Phenotype (6M) & \textemdash & (256, 1, 8, 2) (2) & (256, 2, 4, 1) (2) & \textemdash & (256, 1, 4, 1) (2) \\
OneFlorida+ & Phenotype (12M) & \textemdash & (256, 1, 8, 1) (4) & \textemdash & (128, 2, 4, 2) (2) & (256, 1, 8, 1) (4) \\
OneFlorida+ & Drug Recommendation & (256, 2, 8, 2) (2) & (256, 2, 4, 2) (2) & (256, 2, 8, 4) (3) & (256, 4, 8, 2) (3) & \textemdash \\
\cmidrule(lr){1-7}
\multicolumn{2}{@{}p{1.75in}}{\textit{Mean modal architecture selection (\%)}} & 26.7 & \textbf{40.0} & 33.3 & 33.3 & \textbf{40.0} \\
\midrule
MIMIC-IV & Mortality & \textemdash & \textemdash & \textemdash & \textemdash & (256, 4, 8, 1) (3) \\
MIMIC-IV & Stay $>$ 7d & \textemdash & \textemdash & (256, 2, 4, 2) (2) & \textemdash & (256, 4, 8, 1) (3) \\
MIMIC-IV & Readmission (3M) & \textemdash & \textemdash & \textemdash & (128, 4, 4, 4) (2) & \textemdash \\
MIMIC-IV & Phenotype (6M) & \textemdash & (256, 1, 8, 1) (2) & (256, 2, 4, 2) (2) & (128, 4, 2, 2) (2) & (256, 1, 8, 1) (2) \\
MIMIC-IV & Phenotype (12M) & \textemdash & \textemdash & (256, 2, 8, 4) (2) & \textemdash & (256, 1, 4, 1) (2) \\
MIMIC-IV & Drug Recommendation & \textemdash & (256, 1, 8, 1) (2) & (256, 2, 8, 4) (2) & (128, 4, 8, 8) (2) & (256, 1, 4, 2) (2) \\
\cmidrule(lr){1-7}
\multicolumn{2}{@{}p{1.75in}}{\textit{Mean modal architecture selection (\%)}} & 20.0 & 26.7 & 33.3 & 30.0 & \textbf{43.3} \\
\bottomrule
\end{tabular*}

\par\smallskip
{\footnotesize \textit{Mean modal architecture selection (\%)} summarizes selection consistency across the six tasks. A dash (\textemdash) indicates that all five random seeds selected distinct architectures. Higher values indicate greater consistency across random seeds.\par}
\end{table}

\subsection{Robustness of the cross-hospital prior}

\begin{figure}[H]
\centering
\includegraphics[width=\linewidth,height=0.72\textheight,keepaspectratio]{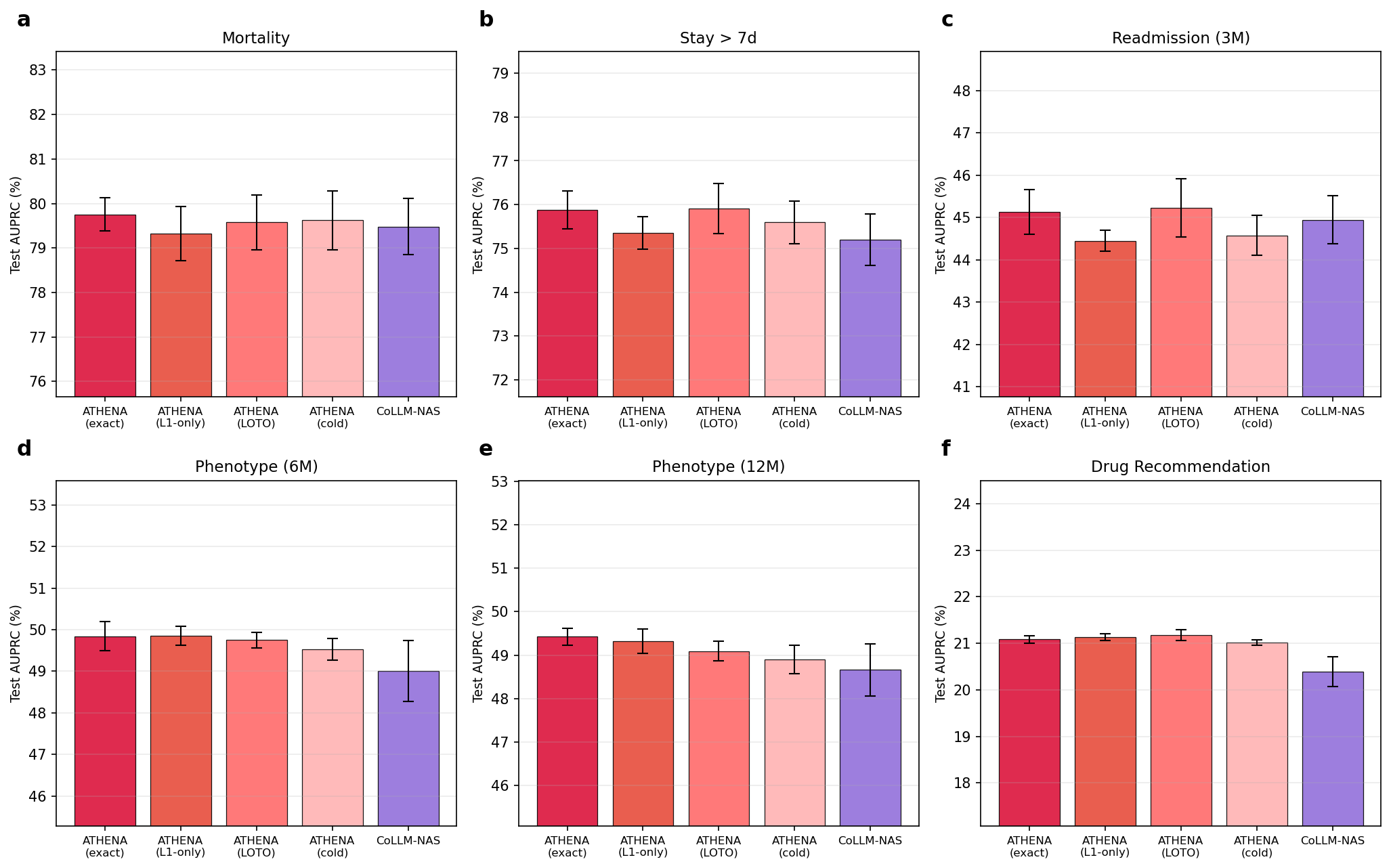}
\caption{Cross-hospital prior ablation on the external MIMIC-IV target. Bars and error bars show the mean and SD, respectively, of test AUPRC across five random seeds for exact task-matched retrieval, \textit{L1-only} retrieval without the \textit{Layer 2} architecture-effect prior, LOTO Retrieval excluding exact task matches, Cold Start without either prior layer, and CoLLM-NAS as the representative baseline.}
\label{fig:ablation-mimic}
\end{figure}

The ablation results separated the contributions of retrieval and architecture-effect guidance (Fig~\ref{fig:ablation-mimic}; \ref{fig:supp-ablation-source}~Fig in S1 Appendix). Removing \emph{Layer 2} (\emph{L1-only}) generally retained competitive performance but reduced AUPRC on several tasks, particularly Phenotype (6M) and Phenotype (12M) on OneFlorida+ and Mortality and Readmission (3M) on MIMIC-IV.

Removing both prior layers (Cold Start) produced the clearest degradation for Phenotype (6M) and Phenotype (12M) on both targets.

When exact task-matched retrieval records were withheld, LOTO Retrieval remained comparable to exact retrieval and was occasionally higher, showing that related-task source records could provide useful fallback guidance.

\section{Discussion}\label{sec_discussion}

In this study, we developed ATHENA, a knowledge-guided agentic NAS framework that combines weight-sharing evaluation with reusable cross-hospital architecture knowledge for Transformer-based EHR modeling. Across six clinical prediction tasks and two target health systems, ATHENA achieved the strongest overall AUPRC ranking across search budgets while requiring substantially less computation than independently pretraining each candidate architecture. ATHENA also produced more consistent architecture selections across repeated searches, and its performance gains were not explained simply by evaluating more candidates or selecting larger models. These findings suggest that architecture-search experience accumulated from prior hospital--task settings can be reused to make model development more efficient and reproducible when adapting EHR models across clinical prediction settings.

Our findings also highlight the complementary roles of the two cross-hospital priors. The retrieval prior provides architecture-level examples from related source settings, whereas the architecture-effect prior summarizes task-specific effects of individual design choices across hospitals. The ablation results support this distinction: removing the architecture-effect prior reduced performance on several tasks, while removing both priors produced larger declines, particularly for multilabel outcomes. Notably, LOTO Retrieval remained competitive even when exact task matches were excluded from the source knowledge base. This result suggests that the transferred information is not limited to reusing architectures previously optimized for the same endpoint; architectural patterns learned from related prediction problems may also provide useful guidance for a new task. At the same time, the observed task-specific architecture effects argue against a single universally optimal Transformer configuration for longitudinal EHR data. Clinical prediction tasks differ in outcome structure, prediction horizon, and the information that must be integrated across a patient's record, and these differences may favor different model capacities and configurations. Cross-hospital transfer of architecture knowledge should therefore complement, rather than replace, adaptation to the target task.

ATHENA extends existing LLM-guided NAS approaches that primarily derive search guidance from information generated during the current search process \cite{zheng2023can,li2026collm}. ATHENA incorporates architecture--performance evidence accumulated at source hospitals and updates this guidance using validation performance at the target site. This distinction is particularly relevant to multi-institutional EHR research, where models are repeatedly developed or adapted across institutions and clinical endpoints. Differences in patient populations, coding practices, clinical care patterns, data density, and outcome prevalence can affect both model performance and the architecture best suited to a prediction problem. Nevertheless, architecture selection is often repeated independently for each new dataset or endpoint. Performing an extensive architecture search for every hospital--task combination can become computationally burdensome, particularly for pretrained Transformer models. Reusing evidence from previous architecture evaluations provides a practical middle ground between applying the same architecture across all settings and repeating a computationally intensive search from scratch for every new study.

The external evaluation on MIMIC-IV further illustrates the potential value of this approach. Architecture knowledge derived from the source hospitals remained informative when transferred to a target health system with different cohort characteristics and clinical data distributions. Importantly, ATHENA did not directly transfer a source architecture as the final model. Instead, source knowledge was used to guide candidate generation, while architecture selection continued to depend on validation performance at the target site. This separation between transferred guidance and local evaluation is important for multi-institutional clinical modeling because an architecture that performs well in one health system may not remain optimal in another because of differences in patient populations, clinical care patterns, and outcome distributions. Architecture transfer may therefore be most useful when prior evidence narrows and informs the search while target-site data retain the final role in model selection.

Several limitations should be considered. First, the evaluation included two target health systems and six downstream prediction tasks. Additional institutions, clinical specialties, coding systems, patient populations, and prediction horizons will be needed to determine how broadly the observed transfer patterns generalize. Although MIMIC-IV provided an external target health system, it represents a single academic medical center and does not capture the heterogeneity of potential real-world deployment environments. Second, the search space was limited to Transformer embedding dimension, depth, number of attention heads, and MLP expansion ratio. Other design choices, including temporal representations, attention mechanisms, prediction heads, tokenization strategies, and training hyperparameters, were held fixed and may interact with the architectural components examined here. Third, although supernet-based evaluation generally preserved architecture rankings, ranking fidelity was lower for some binary outcomes, particularly Readmission (3M), indicating that the accuracy of weight-inherited subnetwork evaluation may vary by prediction task. Fourth, the reported target-search cost does not include the upfront computation required to construct the source-site architecture knowledge base. The efficiency advantage of ATHENA therefore depends on reuse: its potential benefit increases when accumulated architecture evidence can support multiple subsequent modeling studies rather than a single search.

In conclusion, ATHENA demonstrates a framework for reusing architecture-search knowledge across hospitals and clinical prediction tasks while preserving local model selection. By combining retrieved architecture examples, architecture-effect guidance, target-site validation, and weight-sharing evaluation, ATHENA identified strong Transformer configurations under limited search budgets while retaining task- and site-specific adaptation. More broadly, this approach reframes architecture search as knowledge that can accumulate across EHR modeling studies rather than a computational process that must be repeated independently for each new prediction problem. Future work should evaluate this strategy across more diverse health systems and model families, expand the range of transferable architectural and training information, and examine whether more efficient and reproducible architecture development can support robust external validation and prospective evaluation of EHR prediction models.

\section*{Acknowledgments}
OpenAI Codex assisted with grammar checking, language polishing, and conversion to the PLOS LaTeX template.

\section*{Author contributions}
\noindent\textbf{Deyi Li:} Conceptualization, Data curation, Formal analysis, Investigation, Methodology, Software, Visualization, Writing -- original draft.\par
\noindent\textbf{Qi Xu:} Conceptualization, Data curation, Formal analysis, Investigation, Methodology, Writing -- original draft.\par
\noindent\textbf{Lingyao Li:} Validation, Writing -- original draft.\par
\noindent\textbf{Tiansheng Wang:} Validation, Writing -- original draft.\par
\noindent\textbf{Muxuan Liang:} Validation, Writing -- original draft.\par
\noindent\textbf{Mei Liu:} Conceptualization, Funding acquisition, Project administration, Resources, Supervision, Validation, Writing -- original draft.\par

\section*{Competing interests}

The authors declare that they have no known competing financial interests or personal relationships that could have appeared to influence the work reported in this paper.

\section*{Code availability}
Code is publicly available at \url{https://github.com/GatorAIM/ATHENA}.

\section*{Data availability}

The patient-level OneFlorida+ data used in this study are not publicly available because of privacy, institutional, and data-use restrictions. Researchers may request access through the OneFlorida+ Clinical Research Network Front Door or Coordinating Center (\url{https://onefl.net/front-door/}); access is contingent on eligibility, network and institutional review, applicable data-use agreements, and ethical or institutional review board approvals. MIMIC-IV is a de-identified, credentialed-access resource available through PhysioNet (\url{https://physionet.org/}). Access requires PhysioNet credentialing, completion of the required research training, and acceptance of the applicable data-use agreement. The study-specific analytic datasets derived from OneFlorida+ and MIMIC-IV cannot be redistributed by the authors; eligible researchers must obtain the source data independently under the respective data-governance requirements.



\section*{Supporting information}
\paragraph*{S1 Appendix. Supplementary methods, tables, and figures.}
Method~\ref{supp:method-cohort} in S1 Appendix and Method~\ref{supp:method-implementation} in S1 Appendix describe cohort eligibility and implementation details; \ref{tab:supp-outcome-defs}~Table in S1 Appendix, \ref{tab:cohort-baseline}~Table in S1 Appendix, \ref{tab:interaction-rules}~Table in S1 Appendix, and \ref{tab:auroc}~Table in S1 Appendix report task definitions, cohort characteristics, architecture interaction rules, and AUROC results; \ref{fig:supp-prior}~Fig in S1 Appendix, \ref{fig:supp-search-trajectory-mimic}~Fig in S1 Appendix, \ref{fig:supp-pareto-mimic}~Fig in S1 Appendix, and \ref{fig:supp-ablation-source}~Fig in S1 Appendix show the architecture-effect prior, MIMIC-IV search trajectories and Pareto analysis, and OneFlorida+ ablations.

\clearpage
\appendix
\begingroup
\setcounter{section}{0}
\setcounter{figure}{0}
\setcounter{table}{0}
\renewcommand{\thesection}{S\arabic{section}}
\renewcommand{\thefigure}{S\arabic{figure}}
\renewcommand{\thetable}{S\arabic{table}}
\captionsetup[figure]{labelformat=supplementary}
\captionsetup[table]{labelformat=supplementary}
\renewcommand{\theHsection}{supp.\arabic{section}}
\renewcommand{\theHfigure}{supp.\arabic{figure}}
\renewcommand{\theHtable}{supp.\arabic{table}}
\titleformat{\section}{\normalsize\bfseries}{Method~\thesection.}{0.5em}{}
\titlespacing*{\section}{0pt}{1.5\baselineskip}{0.6\baselineskip}
\setlength{\parindent}{0pt}
\setlength{\parskip}{0.6\baselineskip}
\makeatletter
\setlength{\@fptop}{0pt}
\makeatother
\section*{S1 Appendix. Supplementary materials}\label{app:supplementary}
\input{appendix.tex}

\endgroup
\end{document}

%% file: appendix.tex
\section{Cohort construction, inclusion criteria, and exclusion criteria}\label{supp:method-cohort}

The following criteria were applied to all cohorts.

\begin{itemize}

\item \textbf{Multimodal completeness.} An admission was included if it had at least one record in each of the diagnosis, medication,
procedure, and laboratory tables. Admissions missing any of these modalities were excluded.

\item \textbf{Age.} Only adult admissions ($\geq 18$ years) were included. Ages of 89 years and older were coded as
90 years and discretized into 20 equal-width bins.

\item \textbf{Number of clinical events.} No minimum number of clinical events was required beyond the multimodal completeness criterion.

\item \textbf{Number of visits.} No minimum number of admissions per patient was required. To limit computational burden, patients with more than eight distinct admissions were excluded.

\end{itemize}

In addition, the following task-specific criteria were applied.

\begin{itemize}

\item \textbf{Readmission follow-up.} Negative labels required 90 days of follow-up. Follow-up was censored at death or the end of available data; readmissions occurring outside MIMIC-IV or the contributing OneFlorida+ site were not observable.

\item \textbf{Drug Recommendation eligibility.} Admissions were included if they contained at least one medication belonging to the 55 retained Anatomical Therapeutic Chemical (ATC) level-4 classes (prevalence $\geq 1\%$). The resulting drug-eligible cohort was constructed independently of the partition into pretraining and downstream pools.

\end{itemize}

\clearpage

\section{Implementation details}\label{supp:method-implementation}

\begin{itemize}

\item \textbf{Architecture constraints and output heads.}
The target-specific parameter limit was $P_{\max}=4{,}000{,}000$. The componentwise maximal
supernet configuration was $(d,L,\gamma,h)=(256,8,8,8)$, and the internal query, key, and value width
was fixed at 256 for every subnet. All search methods operated on the legal space
$\mathcal{A}_{H,t}$ defined in the main-text Methods. In the shared-supernet comparison, all methods used the same hospital-specific pretrained
supernet.

\item \textbf{Cross-hospital prior.}
The source prior used four de-identified OneFlorida+ sites and 100 distinct architectures per
source-hospital--task pair; both target cohorts were excluded from prior construction. Source metrics
were averaged over the three checkpoints with the highest source-validation area under the precision--recall curve (AUPRC). ATHENA retrieved
$K=5$ source architectures from the task-matched or fallback source set. In the task descriptor,
$\widetilde{C}_t=\log(C_t)/\log(20)$ and $\widetilde{\Delta}_t=\Delta_t/365$, with $\Delta_t=0$ for
same-admission outcomes.
Label entropy and prevalence in the task descriptor are computed on the training split only,
for both source and target hospitals; validation and test labels are never used for descriptor construction.
Validation labels are used separately for candidate evaluation and model selection. Architecture-effect labels
and interaction rules used normal-approximation 95\% confidence intervals (CIs), and interaction cells required at least
three records.

\item \textbf{Agentic search.}
ATHENA permitted at most $R=3$ critique passes per search round and terminated after
$F_{\max}=3$ consecutive rounds without an accepted legal candidate. The maximum evaluation budget
was $B=30$. Results were summarized after 5, 20, and 30 evaluations.

\item \textbf{Masked language modeling (MLM) and supernet pretraining.}
For MLM, 15\% of non-special tokens were selected; 80\% of selected tokens were replaced by
\mbox{\texttt{[MASK]}}, 10\% by a random token of the same type, and 10\% were unchanged.
Supernet pretraining ran for at most 100 epochs with validation-loss patience 5. Optimization used
AdamW, batch size 64, learning rate $2\times10^{-4}$, weight decay $10^{-2}$, and gradient clipping
at norm 1.0. The learning-rate schedule used linear warm-up over the first 10\% of epochs followed by
cosine decay to 1\% of the peak rate.

\item \textbf{Candidate fine-tuning and checkpoint selection.}
Candidate fine-tuning used the same optimizer settings for at most 30 epochs with validation-AUPRC
patience 5. Search-time metrics were averaged over the three epochs with the highest validation AUPRC,
whereas the weights from the single best validation-AUPRC epoch were retained for final test evaluation.
Attention dropout, residual dropout, and drop-path rate were each 0.1.

\item \textbf{Large language model (LLM) configuration.}\\
All LLM-based methods used \texttt{anthropic/claude-3.5-haiku}. The \emph{Proposal}, \emph{Critic}, and Strategy
modules returned structured output in JavaScript Object Notation (JSON); invalid responses were retried and otherwise contributed to the
consecutive-failure criterion.
\end{itemize}

\clearpage

\begin{table}[p]
\caption{Definitions of the six clinical prediction tasks. Each task is defined relative to the
\emph{target admission}, i.e., the hospital admission being modeled. ACE denotes angiotensin-converting enzyme; NSAIDs denote nonsteroidal anti-inflammatory drugs.}
\label{tab:supp-outcome-defs}
\footnotesize
\begin{tabular*}{\tblwidth}{@{}>{\raggedright\arraybackslash}p{\tblwidth}@{}}
\toprule

\textbf{Mortality (binary):}
Whether the patient dies during the target admission ($1$ if a recorded in-hospital death time is
present, $0$ otherwise). The outcome is assessed only during the target admission, from admission
to discharge or death, with no post-discharge follow-up. \\

\midrule

\textbf{Stay $>$ 7d (binary):}
Whether the length of stay of the target admission exceeds seven days ($1$ if $>7$ days, $0$
otherwise), measured from admission to discharge. The outcome is defined for the single target
admission rather than as a cumulative measure across admissions, and all admissions are included
regardless of discharge disposition. \\

\midrule

\textbf{Readmission (3M) (binary):}
Whether the patient has a subsequent hospital admission within 3 months (90 days) of the target
admission ($1$ if a subsequent admission occurs within 90 days, $0$ otherwise). The interval is
measured from the discharge date of the target admission to the admission date of the subsequent
admission. \\

\midrule

\textbf{Phenotype (6M) (multilabel, 18 classes):}
The 18 predefined phenotype categories present among the diagnoses recorded at the patient's next
hospital admission, provided that the next admission occurs within 6 months (180 days) of the
target admission. Only target admissions followed by a qualifying next admission are included.

The 18 categories are: (1) acute and unspecified renal failure; (2) acute cerebrovascular disease;
(3) acute myocardial infarction; (4) cardiac dysrhythmias; (5) chronic kidney disease;
(6) chronic obstructive pulmonary disease; (7) conduction disorders; (8) congestive heart failure
(nonhypertensive); (9) coronary atherosclerosis and related; (10) disorders of lipid metabolism;
(11) essential hypertension; (12) fluid and electrolyte disorders; (13) gastrointestinal
hemorrhage; (14) hypertension with complications; (15) other liver diseases; (16) other lower
respiratory disease; (17) pneumonia; and (18) septicemia (except in labor). \\

\midrule

\textbf{Phenotype (12M) (multilabel, 18 classes):}
The same 18 phenotype categories are used as in the 6-month task, but the next hospital admission
must occur within 12 months (365 days) of the target admission. The label is an 18-dimensional
binary vector, and only target admissions followed by a qualifying next admission are included. \\

\midrule

\textbf{Drug Recommendation (multilabel, 55 classes):}
The medication classes prescribed during the target admission, based on the admission diagnoses
and the patient's prior visit history. Ground truth consists of physician-prescribed drugs mapped
to Anatomical Therapeutic Chemical (ATC) level-4 classes. Of the 140 ATC level-4 classes, the
55 classes present in $\geq 1\%$ across admissions (prevalence, 1.0--21.8\%) are retained. The
resulting label is a 55-dimensional binary vector.
For model input, diagnoses from the target admission are retained, whereas medication, laboratory,
and procedure tokens from that admission are masked; all tokens from prior visits remain available.

The 55 ATC level-4 classes are: (1) antithrombotic agents (B01A); (2) other analgesics and
antipyretics (N02B); (3) stomatological preparations (A01A); (4) laxatives (A06A); (5) drugs for
peptic ulcer and gastro-oesophageal reflux (A02B); (6) other mineral supplements (A12C);
(7) lipid-modifying agents (C10A); (8) irrigating solutions (B05C); (9) calcium (A12A);
(10) beta-blocking agents (C07A); (11) opioids (N02A); (12) antidepressants (N06A);
(13) antiemetics and antinauseants (A04A); (14) ACE inhibitors, plain (C09A); (15) antacids (A02A);
(16) thyroid preparations (H03A); (17) pancreatic hormones (H04A); (18) insulins and analogues
(A10A); (19) selective calcium-channel blockers (C08C); (20) other beta-lactam antibacterials
(J01D); (21) high-ceiling diuretics (C03C); (22) vitamin B12 and folic acid (B03B);
(23) intestinal antiinfectives (A07A); (24) anxiolytics (N05B); (25) adrenergics, inhalants
(R03A); (26) antiepileptics (N03A); (27) other cardiac preparations (C01E); (28) thiazide
diuretics (C03A); (29) cardiac vasodilators (C01D); (30) topical antihemorrhoidals (C05A);
(31) class I and III antiarrhythmics (C01B); (32) angiotensin-II receptor blockers (C09C);
(33) quinolone antibacterials (J01M); (34) antipsychotics (N05A); (35) antigout preparations
(M04A); (36) potassium (A12B); (37) intestinal anti-inflammatory agents (A07E); (38) penicillins
(J01C); (39) general anesthetics (N01A); (40) arteriolar smooth-muscle agents (C02D);
(41) anti-inflammatory and antirheumatic NSAIDs (M01A); (42) dermatological corticosteroids
(D07A); (43) propulsives (A03F); (44) antipruritics (D04A); (45) hypnotics and sedatives (N05C);
(46) other dermatological preparations (D11A); (47) drugs used in addictive disorders (N07B);
(48) cardiac stimulants, excluding glycosides (C01C); (49) topical nasal decongestants (R01A);
(50) potassium-sparing agents (C03D); (51) direct-acting antivirals (J05A); (52) sulfonamides
and trimethoprim (J01E); (53) cough suppressants (R05D); (54) bacterial vaccines (J07A); and
(55) blood glucose-lowering drugs, excluding insulins (A10B). \\

\bottomrule
\end{tabular*}
\end{table}

\clearpage
\begin{table}[p]
\caption{Baseline characteristics of the six study cohorts. Sites A--D are OneFlorida+ source sites used to construct the cross-hospital architecture prior, Site E is the held-out OneFlorida+ internal target, and MIMIC-IV is the external target. Continuous variables are reported as mean (SD) or median [IQR] and categorical variables as n (\%). SD denotes standard deviation and IQR denotes interquartile range. Comorbidities are reported as the number and percentage of admissions with the corresponding diagnosis.}
\label{tab:cohort-baseline}
\footnotesize
\setlength{\tabcolsep}{3.5pt}
\renewcommand{\arraystretch}{1.12}

\begin{tabular*}{\tblwidth}{@{\extracolsep{\fill}}
>{\raggedright\arraybackslash}p{3.5cm}
*{6}{>{\centering\arraybackslash}p{1.82cm}}
@{}}

\toprule
& \multicolumn{4}{c}{\textit{OneFlorida+ Source Sites}}
& \shortstack{\textit{Internal}\\\textit{Target}}
& \shortstack{\textit{External}\\\textit{Target}} \\
\cmidrule(lr){2-5}
\textbf{Characteristic}
& \textbf{Site A} & \textbf{Site B} & \textbf{Site C}
& \textbf{Site D} & \textbf{Site E} & \textbf{MIMIC-IV} \\
\midrule

\multicolumn{7}{@{}l}{\textit{Demographics}}\\

Patients, n
& 52,831 & 55,735 & 88,527 & 58,282 & 84,479 & 60,706 \\

Admissions, n
& 144,313 & 119,579 & 200,911 & 97,303 & 135,955 & 84,492 \\

Female, n (\%)
& 25,428 (48.13) & 27,594 (49.51) & 39,775 (44.93) & 27,952 (47.96) & 46,565 (55.12) & 30,171 (49.7) \\

\addlinespace
\multicolumn{7}{@{}l}{\textit{Utilization}}\\

Admissions per patient, mean (SD)
& 2.7 (1.7) & 2.1 (1.5) & 2.3 (1.4) & 1.7 (1.1) & 1.6 (1.1) & 1.4 (0.9) \\

Length of stay, days, median [IQR]
& 2 [0--4] & 4 [1--7] & 4 [2--7] & 5 [3--16] & 4 [2--7] & 4 [2--8] \\

Diagnoses per admission, mean (SD)
& 10.8 (7.7) & 14.2 (10.6) & 19.1 (15.6) & 12.5 (8.3) & 5.4 (4.2) & 11.9 (7.2) \\

Medications per admission, mean (SD)
& 11.3 (7.1) & 3.5 (2.7) & 13.1 (9.9) & 3.2 (2.3) & 3.9 (2.8) & 3.3 (2.5) \\

\addlinespace
\multicolumn{7}{@{}l}{\textit{Outcomes}}\\

In-hospital mortality, n (\%)
& 644 (0.4) & 1,263 (1.1) & 4,955 (2.5) & 2,226 (2.3) & 5,317 (3.9) & 3,577 (4.2) \\

Stay $>$ 7d, n (\%)
& 13,790 (9.6) & 26,063 (21.8) & 30,841 (15.4) & 33,049 (34.0) & 28,317 (20.8) & 23,031 (27.3) \\

Readmission (3 month), n (\%)
& 30,264 (21.0) & 24,455 (20.5) & 47,990 (23.9) & 19,495 (20.0) & 23,242 (17.1) & 10,727 (12.7) \\

\addlinespace
\multicolumn{7}{@{}l}{\textit{Comorbidities (n, \% of admissions)}}\\

Essential hypertension
& 50,545 (35.0) & 52,639 (44.0) & 85,260 (42.4) & 33,248 (34.2) & 26,591 (19.6) & 35,443 (41.9) \\

Disorders of lipid metabolism
& 50,156 (34.8) & 33,518 (28.0) & 55,822 (27.8) & 23,269 (23.9) & 5,368 (3.9) & 30,498 (36.1) \\

Cardiac dysrhythmias
& 26,555 (18.4) & 20,763 (17.4) & 37,695 (18.8) & 12,110 (12.4) & 7,644 (5.6) & 19,681 (23.3) \\

Congestive heart failure
& 20,014 (13.9) & 19,278 (16.1) & 28,832 (14.4) & 10,740 (11.0) & 6,979 (5.1) & 13,789 (16.3) \\

Coronary atherosclerosis
& 8,808 (6.1) & 4,809 (4.0) & 9,244 (4.6) & 3,055 (3.1) & 1,543 (1.1) & 19,675 (23.3) \\

Acute/unspecified renal failure
& 18,337 (12.7) & 18,995 (15.9) & 29,141 (14.5) & 12,409 (12.8) & 12,003 (8.8) & 13,861 (16.4) \\

Chronic kidney disease
& 10,894 (7.5) & 9,064 (7.6) & 15,732 (7.8) & 6,697 (6.9) & 3,791 (2.8) & 13,489 (16.0) \\

Chronic obstructive pulmonary disease
& 11,814 (8.2) & 11,984 (10.0) & 20,920 (10.4) & 4,581 (4.7) & 3,957 (2.9) & 7,634 (9.0) \\

Pneumonia
& 6,468 (4.5) & 7,449 (6.2) & 15,576 (7.8) & 4,235 (4.4) & 6,086 (4.5) & 5,971 (7.1) \\

Fluid and electrolyte disorders
& 29,744 (20.6) & 36,240 (30.3) & 49,495 (24.6) & 29,854 (30.7) & 13,787 (10.1) & 19,550 (23.1) \\

\bottomrule
\end{tabular*}
\end{table}
\clearpage
\begin{table}[H]
\caption{\textit{Layer 2} pairwise architecture-effect rules supplied to the search prompt ($\mathcal{P}_{\mathrm{meta}}$). For each task, the interaction model uses the two most influential architecture features. The table reports feature-level combinations with estimated effects whose 95\% confidence intervals (CIs) exclude zero.}
\label{tab:interaction-rules}
\footnotesize
\setlength{\tabcolsep}{4pt}
\renewcommand{\arraystretch}{1.25}
\begin{tabular*}{\tblwidth}{@{\extracolsep{\fill}}>{\raggedright\arraybackslash}p{3.0cm}>{\raggedright\arraybackslash}p{3.0cm}>{\raggedright\arraybackslash}p{8.4cm}@{}}
\toprule
\textbf{Task} & \textbf{Feature pair} & \textbf{Rule supplied to the prompt} \\
\midrule
Mortality
& \texttt{embed\_dim} $\times$ \texttt{heads}
& Prefer \texttt{embed\_dim} $\in \{128,256\}$, with the preference for 256 applying when \texttt{heads} $\geq 2$; discourage \texttt{embed\_dim} $\in \{32,64\}$. \\

Stay $>$ 7d
& \texttt{depth} $\times$ \texttt{embed\_dim}
& Prefer \texttt{depth} $\geq 2$; discourage \texttt{depth} $=1$ across all \texttt{embed\_dim} levels. \\

Readmission (3M)
& \texttt{heads} $\times$ \texttt{embed\_dim}
& Prefer \texttt{heads} $\in \{4,8\}$; discourage \texttt{heads} $\in \{1,2\}$ across all \texttt{embed\_dim} levels. \\

Phenotype (6M)
& \texttt{embed\_dim} $\times$ \texttt{depth}
& Prefer \texttt{embed\_dim} $\in \{128,256\}$; discourage \texttt{embed\_dim} $\in \{32,64\}$ across all \texttt{depth} levels. \\

Phenotype (12M)
& \texttt{embed\_dim} $\times$ \texttt{depth}
& Prefer \texttt{embed\_dim} $\in \{128,256\}$; discourage \texttt{embed\_dim} $\in \{32,64\}$ across all \texttt{depth} levels. \\

Drug Recommendation
& \texttt{embed\_dim} $\times$ \texttt{depth}
& Prefer \texttt{embed\_dim} $\in \{128,256\}$; discourage \texttt{embed\_dim} $\in \{32,64\}$ across all \texttt{depth} levels. \\
\bottomrule
\end{tabular*}
\end{table}

\begin{table}[H]
\caption{Test area under the receiver operating characteristic curve (AUROC) at a search budget of 30 architecture evaluations (mean $\pm$ standard deviation over five random seeds). Within each hospital--task comparison, the best mean is shown in \textbf{bold} and the second-best mean is \underline{underlined}; ties share the same mark. \textit{Avg. Rank} denotes the mean rank across the 12 hospital--task comparisons (1 = best; ties receive the average rank), with the number of best-in-row comparisons shown in parentheses. EA denotes evolutionary algorithm.}
\label{tab:auroc}

\footnotesize
\setlength{\tabcolsep}{1.3pt}
\renewcommand{\arraystretch}{1.25}
\begin{tabular*}{\tblwidth}{@{\extracolsep{\fill}}>{\raggedright\arraybackslash}p{1.95cm}>{\raggedright\arraybackslash}p{2.1cm}*{5}{>{\centering\arraybackslash}p{2.2cm}}@{}}
\toprule
Target & Task & Random Search & EA & GENIUS & CoLLM-NAS & ATHENA \\
\midrule

OneFlorida+ & Mortality
& 89.15 $\pm$ 0.72
& \textbf{89.71 $\pm$ 0.51}
& 89.07 $\pm$ 1.15
& 88.94 $\pm$ 1.49
& \underline{89.30 $\pm$ 0.95} \\

OneFlorida+ & Stay $>$ 7d
& 90.32 $\pm$ 0.31
& 90.31 $\pm$ 0.13
& \textbf{90.46 $\pm$ 0.13}
& \underline{90.39 $\pm$ 0.30}
& 90.30 $\pm$ 0.25 \\

OneFlorida+ & Readmission (3M)
& 70.22 $\pm$ 0.28
& \textbf{70.52 $\pm$ 0.48}
& 70.38 $\pm$ 0.19
& 70.29 $\pm$ 0.37
& \underline{70.51 $\pm$ 0.35} \\

OneFlorida+ & Phenotype (6M)
& \underline{78.74 $\pm$ 0.55}
& \textbf{78.82 $\pm$ 0.31}
& 78.39 $\pm$ 0.32
& 78.18 $\pm$ 0.71
& \textbf{78.82 $\pm$ 0.25} \\

OneFlorida+ & Phenotype (12M)
& \underline{79.86 $\pm$ 0.40}
& 79.84 $\pm$ 0.44
& 79.64 $\pm$ 0.26
& 78.85 $\pm$ 0.25
& \textbf{80.05 $\pm$ 0.10} \\

OneFlorida+ & Drug Recommendation
& 71.21 $\pm$ 0.08
& 71.26 $\pm$ 0.08
& \textbf{71.31 $\pm$ 0.07}
& 71.20 $\pm$ 0.18
& \underline{71.28 $\pm$ 0.09} \\

\cmidrule(lr){1-7}

MIMIC-IV & Mortality
& 96.09 $\pm$ 0.14
& \underline{96.18 $\pm$ 0.06}
& 96.13 $\pm$ 0.17
& \underline{96.18 $\pm$ 0.16}
& \textbf{96.19 $\pm$ 0.16} \\

MIMIC-IV & Stay $>$ 7d
& 85.14 $\pm$ 0.26
& \underline{85.43 $\pm$ 0.20}
& 85.33 $\pm$ 0.25
& 85.06 $\pm$ 0.36
& \textbf{85.56 $\pm$ 0.39} \\

MIMIC-IV & Readmission (3M)
& 67.89 $\pm$ 0.19
& \underline{67.90 $\pm$ 0.78}
& 67.77 $\pm$ 0.45
& 67.84 $\pm$ 0.23
& \textbf{68.08 $\pm$ 0.40} \\

MIMIC-IV & Phenotype (6M)
& \underline{80.04 $\pm$ 0.47}
& 79.89 $\pm$ 0.48
& 79.83 $\pm$ 0.52
& 79.47 $\pm$ 0.58
& \textbf{80.07 $\pm$ 0.25} \\

MIMIC-IV & Phenotype (12M)
& \underline{79.38 $\pm$ 0.24}
& \underline{79.38 $\pm$ 0.27}
& 79.15 $\pm$ 0.44
& 78.89 $\pm$ 0.61
& \textbf{79.53 $\pm$ 0.19} \\

MIMIC-IV & Drug Recommendation
& 78.25 $\pm$ 0.17
& \textbf{78.37 $\pm$ 0.07}
& 77.90 $\pm$ 0.46
& 77.41 $\pm$ 0.46
& \underline{78.32 $\pm$ 0.17} \\

\cmidrule(lr){1-7}

\multicolumn{2}{@{}l}{\textit{Avg. Rank (best-in-row)}}
& 3.29 (0)
& \underline{2.21 (4)}
& 3.42 (2)
& 4.38 (0)
& \textbf{1.71 (7)} \\

\bottomrule
\end{tabular*}
\end{table}

\clearpage

\clearpage

\begin{figure}[H]
\centering
\includegraphics[width=\textwidth,height=0.78\textheight,keepaspectratio]{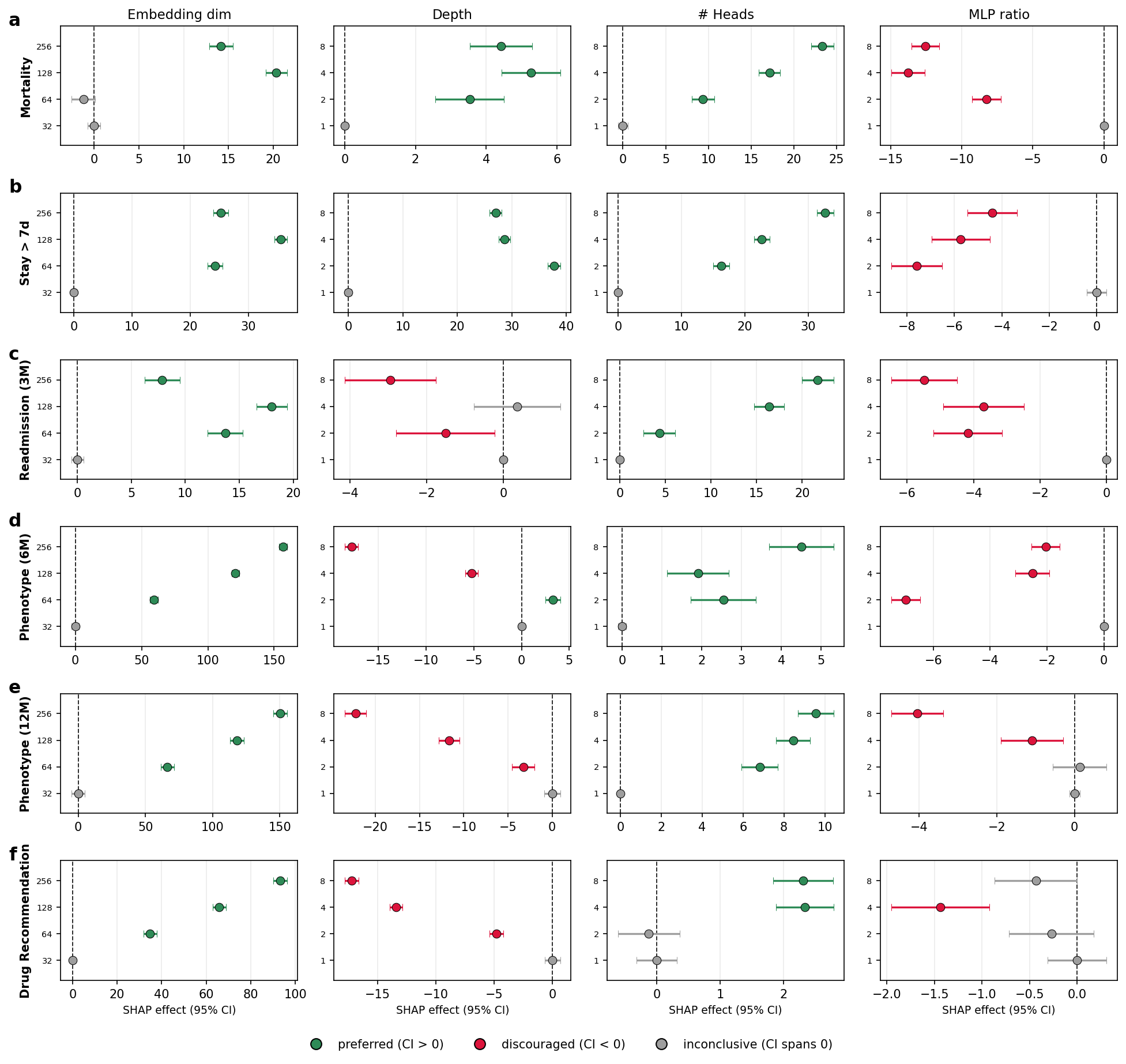}
\caption{Cross-hospital architecture prior learned by the \textit{Layer 2} meta-regression across four OneFlorida+ source sites. Rows denote the six prediction tasks: (\textbf{a}) Mortality, (\textbf{b}) Stay $>$ 7d, (\textbf{c}) Readmission (3M), (\textbf{d}) Phenotype (6M), (\textbf{e}) Phenotype (12M), and (\textbf{f}) Drug Recommendation; columns denote architecture dimensions. Points show mixed-effects SHapley Additive exPlanations (SHAP) estimates with 95\% CIs for each hyperparameter level. Green indicates preferred levels (CI $>$ 0), red indicates discouraged levels (CI $<$ 0), and gray indicates inconclusive levels whose CIs span 0. Gray points at zero without CIs denote reference levels.}
\label{fig:supp-prior}
\end{figure}

\clearpage
\begin{figure}[H]
\centering
\includegraphics[width=\textwidth]{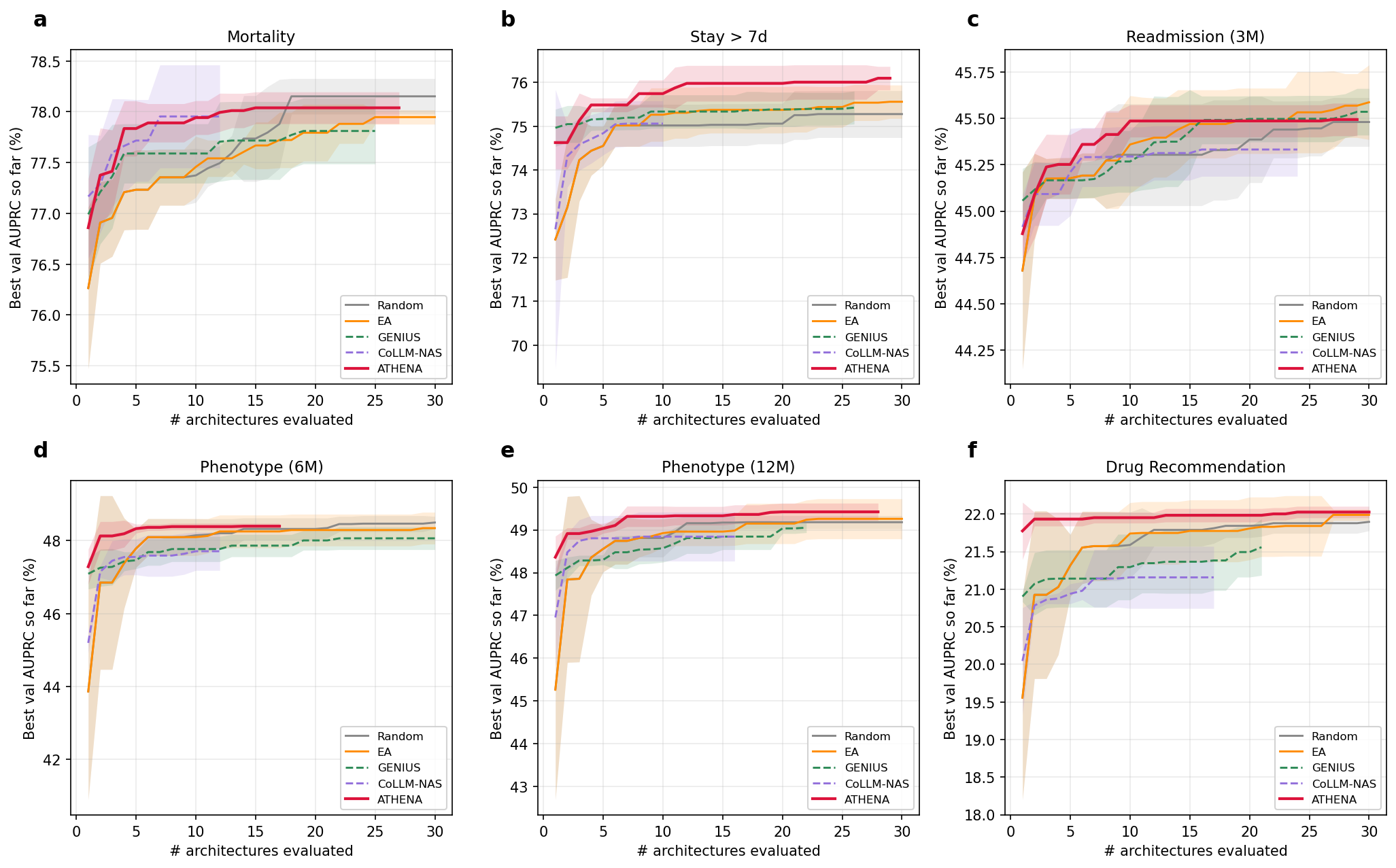}
\caption{Search trajectories on the held-out MIMIC-IV target for (\textbf{a}) Mortality, (\textbf{b}) Stay $>$ 7d, (\textbf{c}) Readmission (3M), (\textbf{d}) Phenotype (6M), (\textbf{e}) Phenotype (12M), and (\textbf{f}) Drug Recommendation. Lines and shaded bands show the mean and SD, respectively, of the best validation AUPRC attained across five random seeds as a function of the number of evaluated architectures. Curves may end before the full budget when proposal saturation yields no new legal architecture.}
\label{fig:supp-search-trajectory-mimic}
\end{figure}

\clearpage
\begin{figure}[H]
\centering
\includegraphics[width=\textwidth]{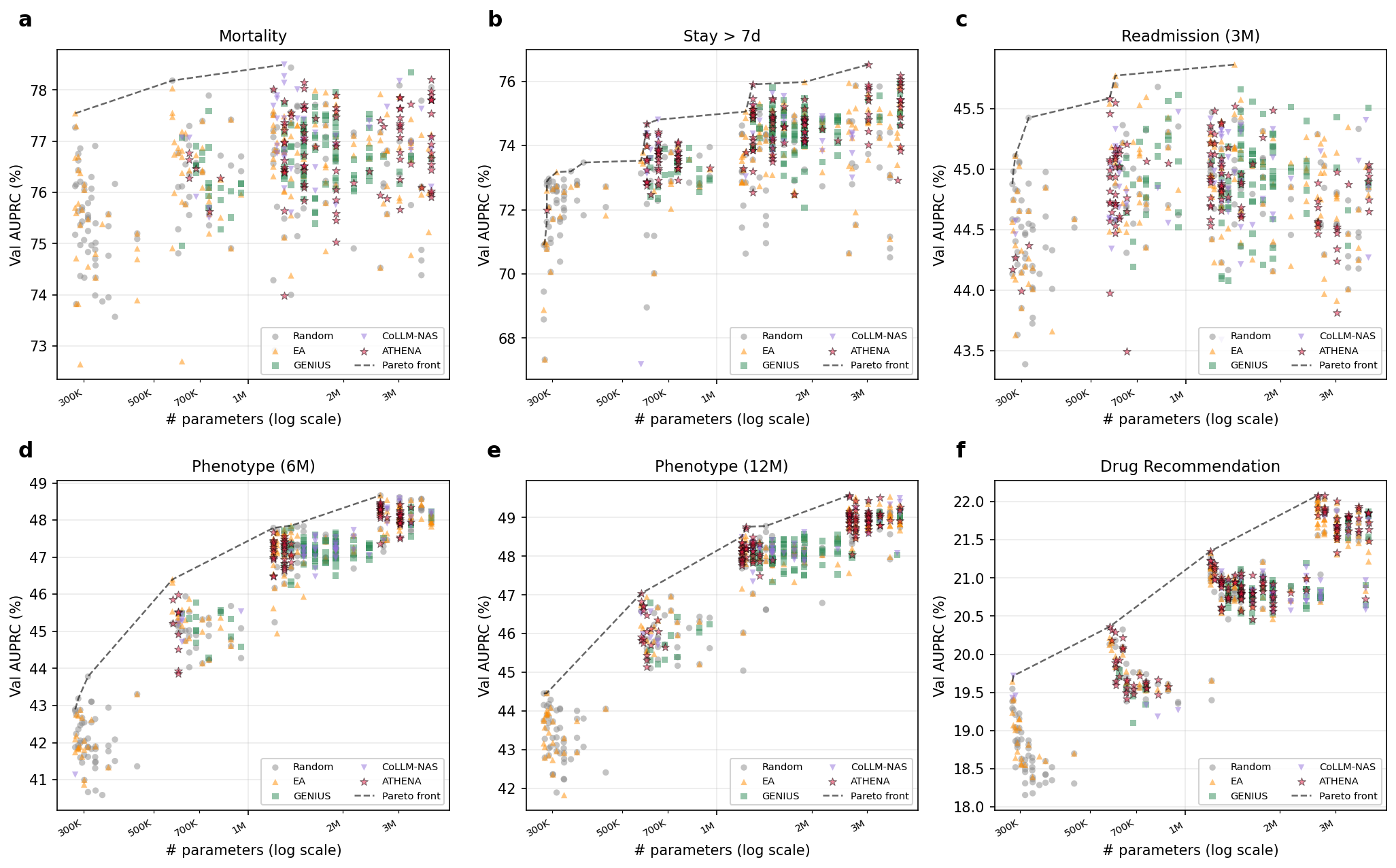}
\caption{Validation AUPRC versus parameter count on the held-out MIMIC-IV target for (\textbf{a}) Mortality, (\textbf{b}) Stay $>$ 7d, (\textbf{c}) Readmission (3M), (\textbf{d}) Phenotype (6M), (\textbf{e}) Phenotype (12M), and (\textbf{f}) Drug Recommendation. Markers denote search methods, and the dashed line indicates the empirical Pareto frontier across evaluated architectures. Validation performance guided architecture search and selection; the selected architecture was evaluated only once on the held-out test set.}
\label{fig:supp-pareto-mimic}
\end{figure}

\clearpage
\begin{figure}[H]
\centering
\includegraphics[width=\textwidth]{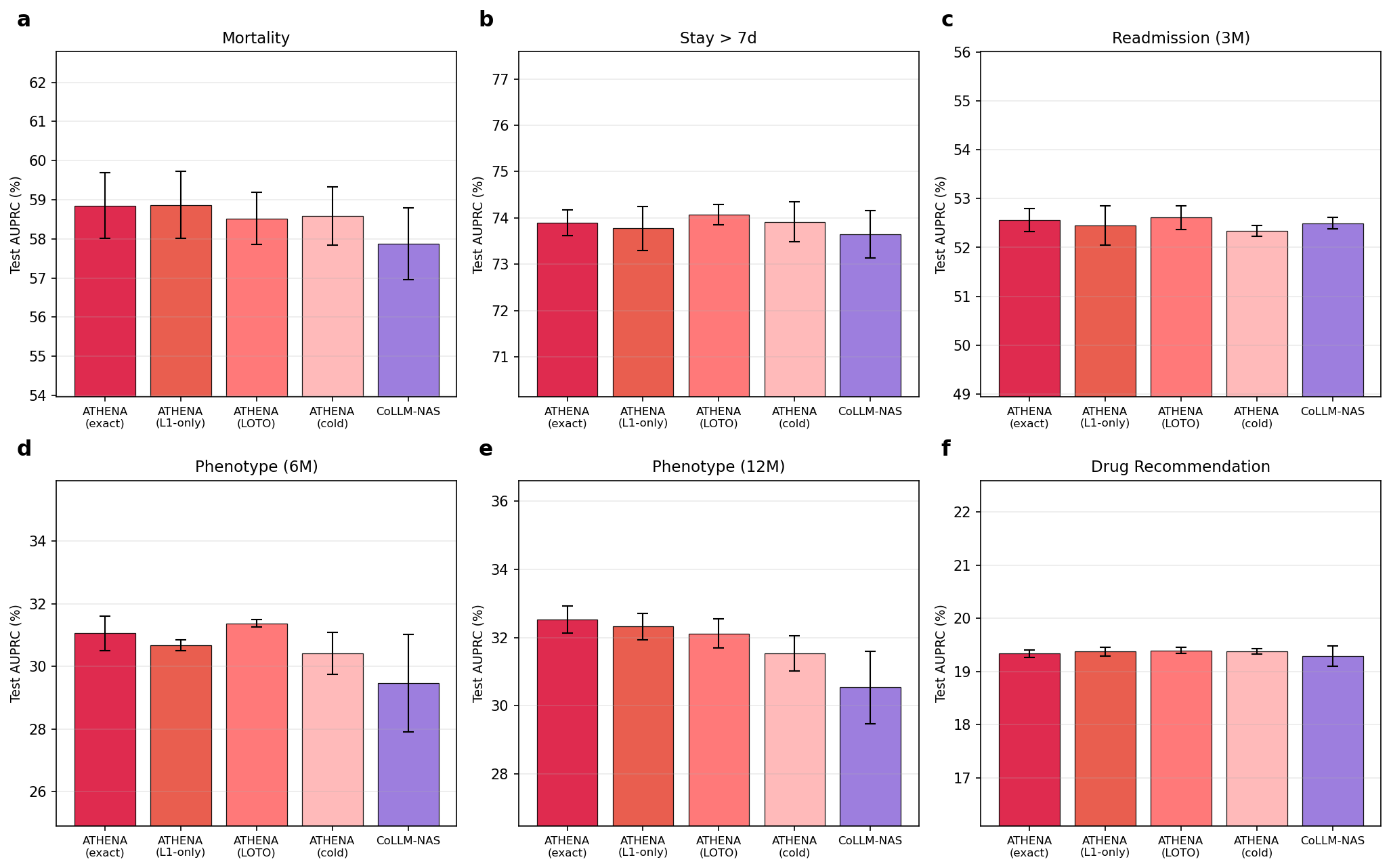}
\caption{Cross-hospital prior ablation on the held-out OneFlorida+ target. Bars and error bars show the mean and SD, respectively, of test AUPRC across five random seeds for exact task-matched retrieval, \textit{L1-only} retrieval without the \textit{Layer 2} architecture-effect prior, Leave-One-Task-Out (LOTO) Retrieval excluding exact task matches, Cold Start without either prior layer, and CoLLM-NAS as the representative baseline.}
\label{fig:supp-ablation-source}
\end{figure}